\documentclass[journal]{IEEEtran}
\usepackage[caption=false,font=normalsize,labelfont=sf,textfont=sf]{subfig}
\usepackage{stfloats}
\usepackage[pdftex]{graphicx}
\usepackage{amsmath}
\usepackage{amssymb}
\usepackage{textcomp,booktabs}
\usepackage[usenames]{color}
\usepackage{colortbl}
\definecolor{mygray}{gray}{.9}
\usepackage{wrapfig}
\usepackage{colortbl,booktabs}
\usepackage{verbatim}
\usepackage[colorlinks,linkcolor=blue]{hyperref}
\usepackage{caption2}

\usepackage{epstopdf}
\usepackage{multirow}
\usepackage{longtable}
\usepackage{rotating}

\begin{document}

\title{Depth Quality Aware Salient Object Detection}

\author{Chenglizhao Chen$^1$
~~~~~~Jipeng Wei$^1$ ~~~~~~Chong Peng$^{1*}$\thanks{Corresponding author: Chong Peng,
pchong1991@163.com. Chenglizhao Chen and Jipeng Wei contributed equally to this
work.}~~~~~~Hong Qin$^2$ \\ ~~~~~~~~~$^1$Qingdao University
~~~~~~~~~~~~~~~~~$^2$Stony Brook University %
%\\ Code\&Data: \url{https://github.com/YunX17/AdaptSaliency}
%\thanks{M. Shell is with the Department
%of Electrical and Computer Engineering, Georgia Institute of Technology, Atlanta,
}

\markboth{IEEE Transactions on Image Processing, VOL.XX, NO.XX, XXX.XXXX}%
{Shell \MakeLowercase{\textit{et al.}}: Bare Demo of IEEEtran.cls for Journals}

\maketitle

\IEEEtitleabstractindextext{
\begin{abstract}
The existing fusion based RGB-D salient object detection methods usually adopt the bi-stream structure to strike the fusion trade-off between RGB and depth (D).
The D quality usually varies from scene to scene, while the SOTA bi-stream approaches are depth quality unaware, which easily result in substantial difficulties in achieving complementary fusion status between RGB and D, leading to poor fusion results in facing of low-quality D.
Thus, this paper attempts to integrate a novel depth quality aware subnet into the classic bi-stream structure, aiming to assess the depth quality before conducting the selective RGB-D fusion.
Compared with the SOTA bi-stream methods, the major highlight of our method is its ability to lessen the importance of those low-quality, no-contribution, or even negative-contribution D regions during the RGB-D fusion, achieving a much improved complementary status between RGB and D.
\end{abstract}


}
\maketitle
\IEEEdisplaynontitleabstractindextext
\IEEEpeerreviewmaketitle

%%%%%%%%%%%%%%%%%%%%%%%%%%%%%%%%%%%%%%%%%%%%%%%%%%%%%%%%%%%%%%%%%%%%
%%%%%%%%%%%%%%%%%%%%%%%%%%%%%%%%%%%%%%%%%%%%%%%%%%%%%%%%%%%%%%%%%%%%

\section{Introduction and Motivation}
The conventional image salient object detection aims to fast locate the most eye attractors, and this topic has received intensive research attentions in recent decades~\cite{fan2018salient}.
As a lightweight pre-processing tool, the down-stream applications of image salient object detection usually include video saliency~\cite{CC2019TIP,CC2017TIP,CC2019TMM2,CC2018TMM,fan2019shifting}, quality assessment~\cite{zhang2015application}, video tracking~\cite{CC2015PR,CC2016PR}, video background extraction~\cite{CC2019CVPR} and so on.

Different to the RGB salient object detection methods which conduct their saliency predictions using RGB information solely~\cite{CC2019TMM1,CC2015TIP}, we focus on the RGB-D salient object detection, which is more challenge than the RGB salient object detection due to the newly available D channel, and we abbreviate ``depth'' as ``D'' for simplicity.
In general, the key rationale of the saliency clue computations in RGB-D images is quite similar to RGB salient object detection methods, in which the RGB-D saliency clues can be easily measured by conducing the contrast computation over the RGB channels and the D channel independently.
Thus, as a subsequent stage after obtaining the RGB-D saliency clues, how to strike an optimal complementary trade-off between RGB and D is the main challenge of the RGB-D salient object detection.

\begin{figure}[t]
\centering
\includegraphics[width=1\linewidth]{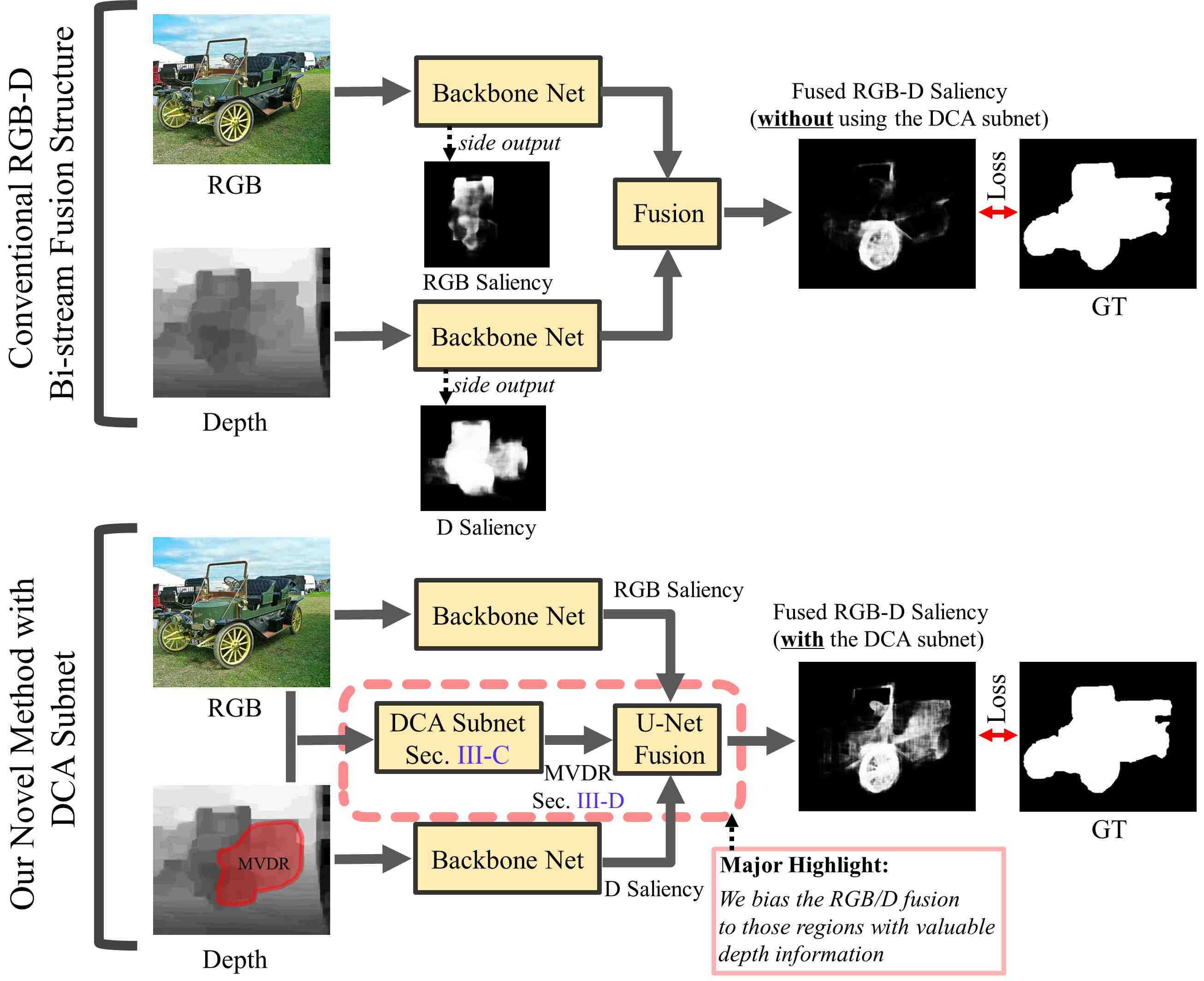} % Reduce the figure size so that it is slightly narrower than the column. Don't use precise values for figure width.This setup will avoid overfull boxes.
\caption{The method pipeline of the proposed method. The major highlight of our method is the newly devised DCA (Depth Contribution Assessment) subnet, which conducts the explicit feature-level RGB-D fusion before performing the selective deep RGB-D fusion, see the blue arrow (more details can be found in Fig.~\ref{fig:Net}).}
\label{fig:motivation}
\end{figure}

In general, the fusion based SOTA (state-of-the-art) RGB-D salient object detection methods~\cite{ren2015exploiting,chen2018progressively,wang2019adaptive,han2018cnns-based,shigematsu2017learning} usually follow the bi-stream structure, in which their two sub-streams compute the RGB saliency clues and the D saliency clues respectively, and these two saliency clues will be latterly combined as the final RGB-D saliency.
After entering the deep learning era, the RGB-D salient object detection methods~\cite{shigematsu2017learning,han2017cnns} widely adopt the pre-trained semantical deep models (e.g., VGG, ResNet) to compute high discriminative deep features automatically for their RGB and D saliency streams, and these deep features will latterly be fed into the fusion layers (e.g., full-connected layers or full-convolutional layers) to achieve the selective deep fusion between RGB and D, see the top section in Fig.~\ref{fig:motivation}.
However, such bi-stream fusion schemes easily reach to their performance bottle-necks due to the following reason:
the depth quality usually varies from scene to scene, while the SOTA bi-stream approaches (e.g.,~\cite{chen2018progressively,wang2019adaptive,han2018cnns-based,shigematsu2017learning}) are depth quality unaware, which easily result in substantial difficulties in achieving complementary fusion status between RGB and D, leading to poor fusion results in facing of low-quality D.

\begin{figure}[t]
\centering
\includegraphics[width=1\linewidth]{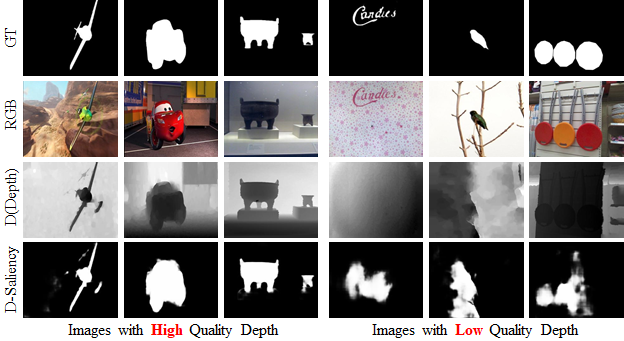} % Reduce the figure size so that it is slightly narrower than the column. Don't use precise values for figure width.This setup will avoid overfull boxes.
\caption{The depth quality demonstrations, in which the left part illustrates those images with high-quality depth, while the right part shows those images with low-quality depth; in general, it is difficult to obtain high performance D Saliency in facing of images with low-quality depth.}
\label{fig:DQualityDemo}
\end{figure}

On the one hand, the D channel is not always capable of benefiting its RGB counterpart; on the contrary, it is merely able to benefit its RGB counterpart occasionally and partially, which suggests us to conduct a biased initial fusion before performing selective deep fusion.
On the other hand, though the widely adopted fusion schemes (e.g., full connected/convolutional layers) are capable of biasing their fusion balances in some extent, yet such biasing degrees are far away from achieving an optimal RGB-D complementary status, because these schemes are unaware of the depth quality during their supervised model training.
For example, the training loss of the D stream still exists when facing a training image with extremely low-quality D (e.g., the 3rd row of the right part in Fig.~\ref{fig:DQualityDemo}), which easily leads to the learning ambiguity or learning over-fitting, preventing the fusion layers to completely bias to the RGB stream.

To handle the above mentioned problems, in this paper, we integrate a novel DCA (Depth Contribution
Assessment) subnet into the bi-stream structure, aiming to conduct depth quality assessment before performing the selective fusion, and we have demonstrated its method overview in the bottom part of Fig.~\ref{fig:motivation}.
Our DCA subnet is inspired by the following two common attributes of those D regions which are capable of benefiting their RGB counterparts during the RGB-D saliency fusion: 1) only those high-quality D regions (e.g., the right part of Fig.~\ref{fig:DQualityDemo}) are potentially able to benefit the RGB stream;
2) among of these high-quality D regions, only a small part of it, which have exhibited different saliency predictions to the RGB saliency stream, are the most valuable D regions during the RGB-D saliency fusion.

Therefore, our method attempts to weakly label those D regions which simultaneously meet the above 2 aspects as the pseudo-GT for the DCA subnet training.
Once the DCA subnet has been trained, we use its predictions to guide the explicit feature-level fusion (i.e., M1-M4 in Fig.~\ref{fig:Net}) before conducting the selective RGB-D deep fusion.
Compared with the conventional selective deep fusion, our novel method is depth quality aware, which is much better at achieving an optimal complementary trade-off between RGB and D.
In summary, the major contributions of this paper can be summarized as follows:
\begin{itemize}
    \item
    We have raised one crucial factor which determines the RGB-D salient object detection performance, i.e., it will be able to achieve a much improved fusion if we lessen the importance of those low-quality, low-contribution, or even negative-contribution D regions beforehand;
	\item
    We have proposed a novel weakly supervised DCA (Depth Contribution Assessment) subnet, aiming to predict which D regions may potentially be able to benefit the RGB stream;
    \item
    We have devised a novel selective fusion network to make full use of the DCA subnet, achieving a much improved complementary fusion status between RGB and D;
    \item
    We have conducted massive quantitative evaluations (i.e., comparisons to 12 most recent SOTA methods over 5 datasets) to validate the effectiveness and show the performance superiority of our method.
    \item
    Both the source code and data are available online at \url{https://github.com/qdu1995/DQSD}, which will be able to benefit the RGB-D salient object detection field.

\end{itemize}

\section{Related Works}
\subsection{The RGB Salient Object Detection Methods}
The RGB image salient object detection methods use RGB information solely to derive image saliency.
The conventional RGB salient object detection methods~\cite{jiang2013salient,peng2016salient,cong2017co,huo2018semisupervised} were developed by using the handcrafted features (e.g., the most representative background prior~\cite{wei2012geodesic}) to conduct the multi-scale contrast/uniqueness computations for the low-level saliency clues, e.g., the classic regional contrast computation~\cite{cheng2014global}.

With the rapid development of deep learning techniques, the SOTA (state-of-the-art) RGB salient object detection deep models~\cite{wang2015deep,lee2016deep,li2018contrast} have outperformed the conventional handcrafted methods in both accuracy and efficiency.
Recently, the research trends of the deep learning based RGB salient object detection mainly include: the multi-scale deep feature integration by using the side-layer output of the low-resolution layers~\cite{DSS}, the stage-wise saliency recurrent~\cite{zhang2017amulet}, the multi-scale spatial/channel attentions~\cite{li2019video}, and the object boundary enforcement~\cite{EGNet}.
Since the RGB salient object detection is not the main foci of this paper, we shall not cover this topic any further.
Also, it is worthy mentioning that the pre-trained RGB saliency deep models can be directly applied as the RGB saliency subnet in the bi-stream network structure.

\subsection{The RGB-D Salient Object Detection Methods}
The RGB-D salient object detection methods can be roughly divided into 2 groups according to their main research focuss: 1) how to reveal saliency clue in D channel; 2) how to fuse RGB saliency with D saliency to achieve an improved RGB-D saliency.
The first group methods mainly follow the conventional handcrafted manner to design novel feature space for an improved D saliency, and their key rationale are quite similar to that of the RGB salient object detection methods.
The most representative method is the work proposed by Feng et al. in~\cite{CVPR_F2016}, which devised a novel depth angular direction based metric to suppress those non-salient areas with high RGB contrasts.

After entering the deep learning era, any off-the-shelf RGB saliency deep models will be able to predict D saliency if we fine-tune it using the D channel instead.
Thus, we shall focus on the RGB-D fusion schemes here.
Qu et al.~\cite{TIP_Q2017} have adopted the conventional handcrafted manners to compute the low-level saliency clues over both RGB and D channels within the superpixel-wise manner, e.g., the local/global contrast computation and the boundary contrast computation.
Then, these low-level saliency clues will be fed into CNN (Convolutional Neural Network) to compute high discriminative deep features (with 4096 dimensions), which will be latterly input into full connected layers to regress towards the RGB-D saliency ground truth, aiming to achieve the selective deep fusion between RGB and D.
Similarly, R.Shigematsu et al.~\cite{shigematsu2017learning} measured D saliency using the newly designed background enclosure distribution, and this novel saliency clue was latterly integrated with multiple top-down and bottom-up saliency clues via CNN framework.
Though much improvements have been made by using such ``hybrid'' none end-to-end methods, it has two major limitations: 1) its fusion performance heavily dependents on the handcrafted low-level saliency clues; 2) its superpixel-wise detection (i.e., the full connected layers) is time consuming in general.

To alleviate the above mentioned limitations, Zhu et al.~\cite{Zhu2018PDNet} have adopted the full convolutional network based bi-stream structure.
As the main stream, its RGB saliency is similar to the SOTA RGB saliency deep models, which follows the classic UNet encoder-decoder structure for a fast RGB saliency computation.
Its major highlight is that the side-layers in the D stream are aligned with the RGB saliency decoder layers, aiming to achieve the feature-level selective RGB-D saliency fusion.
Similarly, Han et al.~\cite{han2017cnns} have followed the bi-stream structure, whose D saliency was computed by using the task-relevant initialization with deep supervision in the hidden layers.
To achieve selective deep fusion, the work~\cite{han2017cnns} has adopted a combination layer to connect its RGB saliency stream to its D saliency stream.
However, due to the lack of shallower semantic information, the RGB-D saliency map predicted by~\cite{han2017cnns} may occasionally exist massive false-alarm detections.

\begin{figure*}
\begin{center}
\includegraphics[width=1\linewidth]{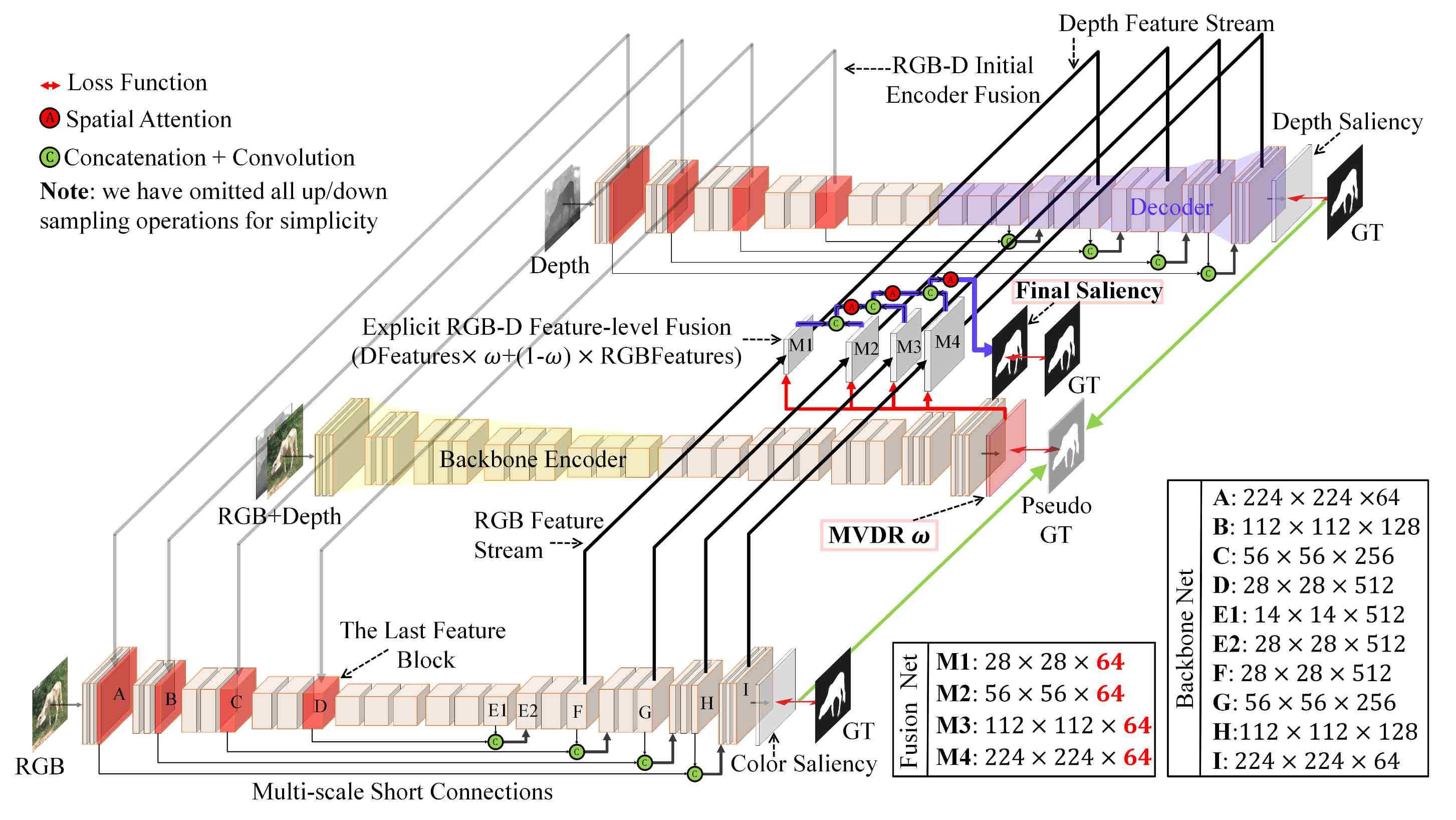}
\end{center}
\captionsetup{justification=centering}
   \caption{The overall network architecture of our proposed method, and we have listed the network details in the bottom-right; we have demonstrated the highlight of this paper in the middle, i.e., the DCA (Depth Contribution Assessment) subnet.}
\label{fig:Net}
\end{figure*}

To increase the information exchange between RGB and D channels, Chen et al.~\cite{chen2018progressively} have devised a residual function to measure the complementary degree between RGB and D, and then it resorted the level-wise supervision to achieve cross-modal RGB-D fusion.
Further, Chen et al.~\cite{chen2019three} have devised a three-stream network, i.e., the conventional RGB-D bi-stream with a novel bottom-up cross-model stream.
The major highlights of this bottom-up cross-modal stream comprise two-fold: 1) this novel stream is able to learn high discriminative deep features; 2) this novel stream is able to complement both RGB and D streams when these two conflicting with each other.
Wang et al.~\cite{wang2019adaptive} have followed the bi-stream network structure to predict RGB saliency and D saliency respectively.
The major highlight of this work is that it has resorted an additional subnet to weakly learn a switch-map, which is formulated by measuring the difference between its RGB saliency and saliency GT, and this switch-map will explicitly guide the RGB-D saliency fusion.
The behind rationale of~\cite{wang2019adaptive} is based on the assumption that those image regions with incorrect RGB saliency may potentially get fixed by using the D channel.
However, its pseudo-GT, which will latterly be used to learn a deep model to predict the RGB-D switch-map, is problematically formulated, because it is almost an impossible task for the deep network to predict which RGB pattern may fail to produce correct saliency detection, resulting in an over-fitted switch-map eventually.

Most recently, Zhao et al.~\cite{zhao2019contrast} have attempted to enhance the depth quality by using the newly designed contrast prior, which transferred the contrast information from the RGB channels to the D channel, aiming to enlarge the depth differences between salient objects and their non-salient surroundings nearby.
In fact, the key motivation of our work is partially similar to~\cite{zhao2019contrast}, i.e., the~\cite{zhao2019contrast} has noted that the RGB-D fusion performance may get degenerated in facing of those low-quality D, thus Zhao et al. has adopted the RGB channels to improve the depth quality from the contrast perspective; in our work, we also attempt to alleviate the side-effect of those low-quality depth information, yet we solve this problem by evaluating the depth quality explicitly before RGB-D fusion, archiving a much improved RGB-D fusion performance.

\section{The Proposed Method}
\subsection{Method Overview}
We have demonstrated the overall network architecture in Fig.~\ref{fig:Net}.
Our method mainly consists of 4 components, including 1) the RGB saliency stream, 2) the D saliency stream, 3) the DCA (Depth Contribution Assessment, Sec.~\ref{sec:DQSN}) subnet and 4) the MSF (Multi-scale Fusion, Sec.~\ref{sec:MSDF}) subnet.

The RGB and D saliency streams adopt the classic UNet~\cite{ronneberger2015u} encoder-decoder network architecture, which recursively makes full use of the multi-scale deep features between in its encoder and decoder layers (Sec.~\ref{sec:SNP}).
Then, based on the stage-wise saliency predictions from the RGB and D saliency streams (i.e., the green arrows in Fig.~\ref{fig:Net}), we formulate the pseudo-GT to weakly train the newly designed DCA subnet (Sec.~\ref{sec:DQSN}), which aims to indicate which regions in the D channel are potentially able to benefit their RGB counterparts for the salient object detection task.
Once the DCA subnet has been trained, we use its prediction ($\omega$) to guide an initial feature-level fusion, which explicitly combines the side-outputs of the RGB saliency stream with the side-outputs of the D saliency stream as the fused RGB-D deep features, i.e., the M1-M4 in Fig.~\ref{fig:Net}.
Moreover, we have devised a novel network to conduct multi-scale selective deep fusion for the previously fused deep features (Sec.~\ref{sec:MSDF}), in which we recursively convolve M1-M4 with spatial attentions to compute high-quality RGB-D salient object detection results (see the blue arrows in Fig.~\ref{fig:Net}).
Specifically, we have connected the RGB encoder layers with the D encoder layers (see the gray arrows in Fig.~\ref{fig:Net}), aiming to ensure a high-quality RGB saliency before performing the subsequent RGB-D fusion.

\subsection{Subnet Preliminaries}
\label{sec:SNP}
In our method, the RGB saliency stream, the D saliency stream and the DCA (Depth Contribution Assessment) subnet are all following the conventional full convolutional network architectures, which respectively take the RGB channel, the D channel and the hybrid RGB+D as input, aiming to regress their individual given input to their learning objectives respectively.
To obtain the discriminative semantical deep features for these subnets, we choose the off-the-shelf VGG19~\cite{simonyan2014very} as the backbone encoder and any other feature backbone (e.g., ResNet~\cite{he2016deep} and Res2Net~\cite{gao2019res2net}) is also OK but it may produce different performances (Table~\ref{tab:addlabel}), see the yellow regions in Fig.~\ref{fig:Net}.
Meanwhile, the decoder layers (the blue regions) take the deepest feature block of the encoder layers as input, which recursively conduct sequential up-samplings ($U$) and convolutions ($Conv$) to ensure an identical output resolution to the input image.
Thus, the overall data flow of the DCA subnet can be represented as Eq.~\ref{eq:DataFlowDepthQ}.
\begin{equation}
\label{eq:DataFlowDepthQ}
\omega = ...Conv\bigg(U\Big(Conv\big(VGG(RGB+D)\big)\Big)\bigg)...,
\end{equation}
where $\omega\in\mathbb{R}^{224\times 224}$ denotes the MVDR (Most Valuable Depth Regions) which are predicted by the DCA subnet; $U$ is the up-sampling operation and $Conv$ is the convolutional operation; $RGBD\in\mathbb{R}^{224\times 224\times 4}$ denotes the $\{RGB+D\}$ input; $VGG$ is the pre-trained VGG19; we use the symbol ``$...$'' to denote those $Conv$ and $U$ operations which are omitted here for simplicity.

To enhance the tiny details of the detected saliency map, the decoder layers in the RGB and D saliency subnets recursively convolve the deep features between consecutive deep layers, which iteratively make full use of the multi-scale deep features.
We represent the detailed dataflow of the RGB saliency subnet as Eq.~\ref{eq:DataFlowRGBD}.
\begin{equation}
\begin{split}
E2 = Conv\Big(C\big(D,U(E1)\big)\Big),\ F = Conv\big(Conv(E2)\big),
\label{eq:DataFlowRGBD}
\end{split}
\end{equation}
where $C$ and $U$ denote the concatenation operation and the up-sampling operation respectively; $D$, $E1$, $E2$ and $F$ can be found in Fig.~\ref{fig:Net}.

\subsection{Depth Contribution Assessment (DCA) Subnet}
\label{sec:DQSN}
As we have mentioned before, the DCA subnet aims to predict which D regions may potentially be able to benefit the RGB saliency subnet for the salient object detection task, and we abbreviate such D regions as ``MVDR (Most Valuable Depth Regions)''.
To train the DCA subnet, we should prepare its training instances (i.e., RGB-D images) with well labelled MVDR as the learning objectives.

In fact, there exists one remarkable common attribute of those MVDR, i.e., the D saliency ($DSal\in\mathbb{R}^{224\times 224}$) of those MVDR should outperform the corresponding RGB saliency ($RGBSal\in\mathbb{R}^{224\times 224}$).
Since the $RGBSal$, the $DSal$, and the human annotated binary saliency $GT$ are simultaneously available for the current training stage, we formulate the pseudo-GT ($pGT=\{P+B\}$) of the DCA training set via Eq.~\ref{eq:P} and Eq.~\ref{eq:B}, attempting to weakly train the DCA subnet for the prediction of those MVDR in the given RGB-D image.

\begin{equation}
\label{eq:P}
{P} = \underset{larger\ DSal\ in \ the\ salient \ regions}{\underbrace{pos(DSal-RGBSal)\odot GT}},
\end{equation}
\begin{equation}
\label{eq:B}
{B} = \underset{smaller\ DSal\ in \ the\ non-salient \ regions}{\underbrace{pos(RGBSal-DSal)\odot (1-GT)}},
\end{equation}
where $GT\in\{0,1\}$ denotes the human well annotated binary saliency ground truth; $\odot$ denotes the element-wise Hadamard product; $pos$ is a function which assigns those negative elements to zero.

Actually, for those regions inside the salient regions indicated by $GT$, Eq.~\ref{eq:P} aims to locate those regions with larger $DSal$ than $RGBSal$, which means that such D regions can better highlight the salient regions than their RGB counterparts.
Meanwhile, for those regions inside the non-salient regions indicated by $GT$, Eq.~\ref{eq:B} aims to locate those regions with smaller $DSal$ than $RGBSal$, which means that such D regions can well suppress those non-salient surroundings.

So, we formulate the $pGT=\{P+B\}$, and its key rationales can be summarized into the following two aspects:\\
1) in the case of those D regions with their $GT=1$, it is quite intuitive to assign these D regions as MVDR if their $DSal>RGBSal$;\\
2) on the other hand, as for those D regions with their $GT=0$, we should also assign these D regions as MVDR if $DSal<RGBSal$.

\begin{figure}[t]
\centering
\includegraphics[width=1\linewidth]{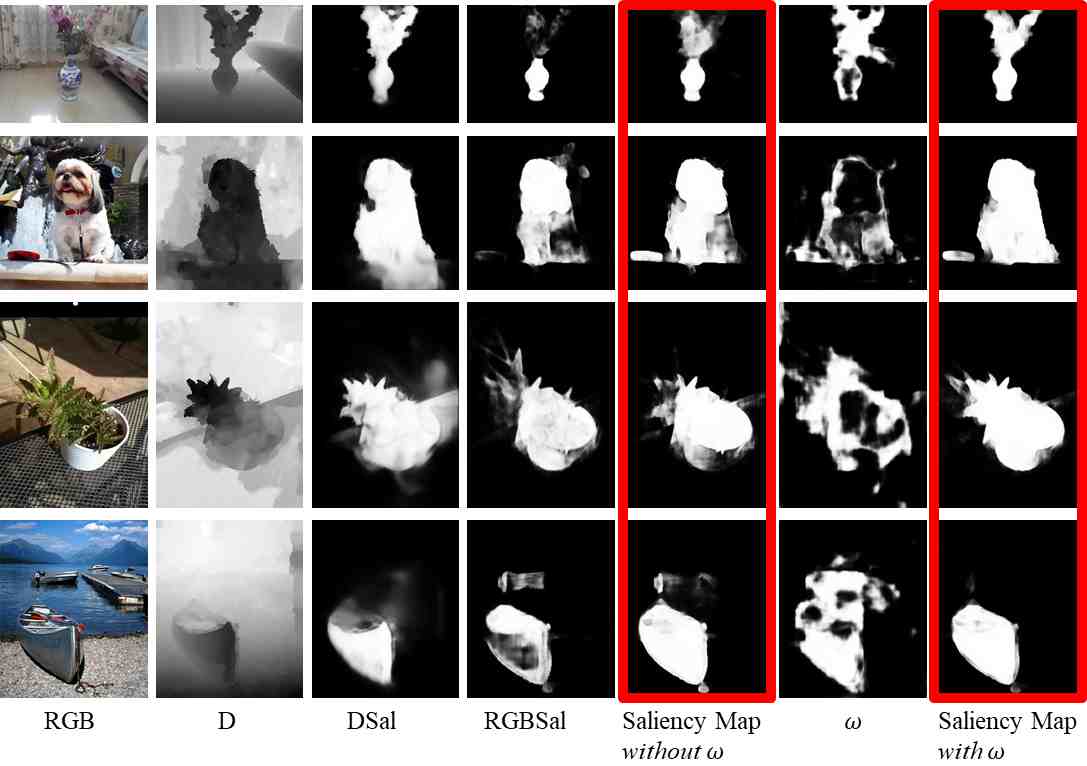} % Reduce the figure size so that it is slightly narrower than the column. Don't use precise values for figure width.This setup will avoid overfull boxes.
\caption{The qualitative demonstrations of the $\omega$ (the 6th column) predicted by DCA subnet, in which the 5th column and the 7th column show the performance variation between these two different fusion schemes, i.e., without using the $\omega$ (5th column) and using $\omega$ (7th column).}
\label{fig:omega}
\end{figure}

Then, we directly use the above weakly formulated pseudo-GTs as the training objective for our DCA subnet, and we represent its learning loss as Eq.~\ref{eq:QNetFlow}.
\begin{equation}
\label{eq:QNetFlow}
Loss = L(\omega, pGT),\  \omega = DeCoder\big(EnCoder(RGBD)\big),
\end{equation}
where $L$ denotes the cross-entropy loss; $\omega\in[0,1]^{224\times 224}$ is the MVDR predicted by the DCA subnet, and we have demonstrated its qualitative results in Fig.~\ref{fig:omega}.

\subsection{Why the MVDR (Most Valuable Depth Regions, $\omega$) Can Be Learned?}
Generally speaking, the key rationale of the deep learning techniques is to memorize all of its given training instances, aiming to generalize toward the given learning objective when facing unseen data.
Thus, a ``reasonable'' learning objective is extremely important for deep networks to output their desired results, e.g., we shouldn't anticipate to train a deep network for an impossible mission, such as predicting lottery winning numbers.

In our case, the learning objective for the DCA (Depth Contribution Assessment) subnet (Sec.~\ref{sec:DQSN}) is to learn how to predict/locate those D regions which are potentially able to benefit their RGB counterparts for the salient object detection task.
So, following the conventional hand-crafted thinking model, we may approximately achieve the aforementioned learning objective by conducting the following two sequential tasks:
1) because only those high-quality D regions may potentially be able to benefit their RGB counterparts, we coarsely locate those high-quality D regions first;
2) then, based on those high-quality D regions determined by the task 1), we preserve those D regions which are able to provide more accurate saliency clues than their RGB counterparts.

Theoretically, it is quite intuitive to fulfill the task 1) via the following principle:
those high-quality D regions must be nearby those image pixels which exhibit strong consistency in their gradient values between the RGB channel and the D channel.
As for the task 2), because both $RGBSal$ and $DSal$ are simultaneously available, we can easily locate the MVDR by simply filtering those D regions which have similar $RGBSal$ and $DSal$.

Thus far, we have provided a feasible hand-crafted solution to approximately full fill the learning objective of the DCA subnet, showing its technical soundness.
Also, it is worthy mentioning that the learned DCA subnet is certain to outperform the aforementioned hand-crafted manner in both accuracy and efficiency.

\subsection{Multi-scale Fusion Subnet}
\label{sec:MSDF}
By using the DCA subnet, we can easily obtain the MVDR (Most Valuable Depth Regions, $\omega$) in the given testing RGB-D image.
Thus, we may simply achieve an explicit RGB-D saliency fusion ($RGBDSal$) by using Eq.~\ref{eq:SimpleFusion}.
\begin{equation}
\label{eq:SimpleFusion}
{RGBDSal} = \omega \odot DSal + (1-\omega) \odot RGBSal.
\end{equation}
In fact, the above hand-crafted $RGBDSal$ is slightly worse than the conventional bi-stream selective fusion, and its fusion performance is heavily dependent on either the $DSal$ or the $RGBSal$.
To further improve, we use the $\omega$ to guide the feature-level selective deep fusion between the RGB saliency subnet and the D saliency subnet.

To be specific, for each one of the last four decoder feature blocks, we convolve it as the side-output with fixed channel number (64) while leaving its resolution unchanged.
For example, as shown in Fig.~\ref{fig:Net}, the first side-output of the RGB saliency subnet ($RGB_{1}^{s}$) can be formulated by using 64 $\{1\times 1\}$ kernels over the feature block $F$.
Thus, we formulate the feature-level fusion as Eq.~\ref{eq:FeatureFusion}.
\begin{equation}
\label{eq:FeatureFusion}
M_i = \omega \odot D_{i}^{s} + (1-\omega) \odot RGB_{i}^{s}, \ i\in[1,4],
\end{equation}
where $M_i$ denote the $i$-th fused RGB-D deep features.

\begin{figure*}
\begin{center}
\includegraphics[width=1\linewidth]{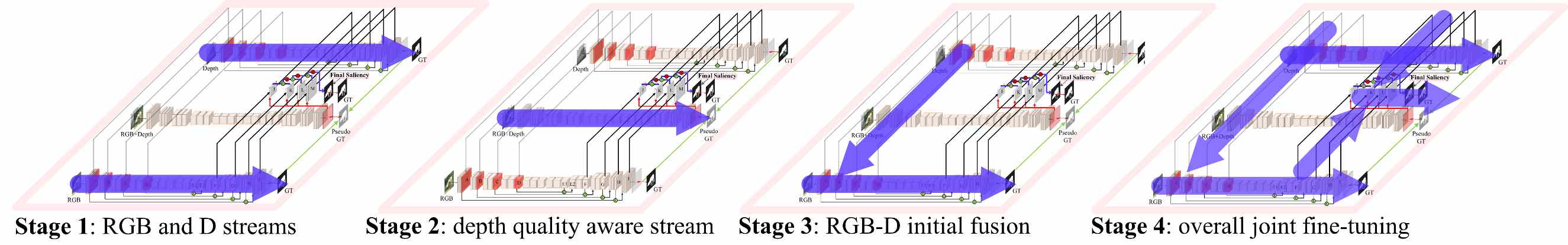}
\end{center}
   \caption{The demonstrations of our stage-wise training scheme (Sec.~\ref{sec:NT}).}
\label{fig:Training}
\end{figure*}

To achieve an optimal complementary fusion status between the multi-scale deep features, i.e., the $M_i$ (Fig.~\ref{fig:Net}), we compute the final RGB-D saliency ($FSal$) via recursive iterations, which consist of the following sequential steps:
\begin{equation}
\begin{split}
{\rm 1.}&\ temp \gets A\Bigg(Conv\Big(C\big(U(M_1),M_2\big)\Big)\Bigg),\\
{\rm 2.}&\ temp \gets A\Bigg(Conv\Big(C\big(U(temp),M_3\big)\Big)\Bigg),\\
{\rm 3.}&\ temp \gets A\Bigg(Conv\Big(C\big(U(temp),M_4\big)\Big)\Bigg),\\
{\rm 4.}&\ FSal \gets Conv(temp),
\label{eq:FinalFlow}
\end{split}
\end{equation}
where $U$ and $C$ respectively denote the up-sampling operation and feature concatenation operation; $A$ denotes the spatial attention operation as Eq.~\ref{eq:A}, in which $h$ denotes a $1\times 1$ convolution with 1 output channel; $temp$ is an auxiliary container.
\begin{equation}
temp \gets \Big(1+Sigmoid\big(h(temp)\big)\Big)\times temp.
\label{eq:A}
\end{equation}

The behind rationale of the above recursive iterations (Eq.~\ref{eq:FinalFlow}) is to make full use of both the salient object localization information provided by the low-resolution layers, and the tiny saliency details provided by the high-resolution layers.
Meanwhile, the spatial attention operations aim to highlight those most valuable saliency clues between different scales.

\subsection{Network Training}
\label{sec:NT}
As shown in Fig.~\ref{fig:Net}, our network consists of 4 components, including: 1) the RGB saliency subnet (top); 2) the D saliency subnet (bottom); 3) the DCA (Depth Contribution Assessment) subnet (middle); 4) the MSF (Multi-scale Fusion) subnet which is above the DCA subnet.

Sine the pseudo-GTs (Sec.~\ref{sec:DQSN}) are indispensable for the DCA subnet training, we first train the RGB and D
saliency subnets respectively (see the stage 1 in Fig.~\ref{fig:Training}).
Next, we weakly train the DCA subnet by using the pseudo-GTs formulated via Eq.~\ref{eq:P} and Eq.~\ref{eq:B}, see the stage 2 in Fig.~\ref{fig:Training}.
Once the DCA subnet has been trained, we use its prediction $\omega$ to guide the feature-level deep fusion, obtaining the multi-scale RGB-D deep features (i.e., M1, M2, M3 and M4 in Fig.~\ref{fig:Training}).
Because the performance of such feature-level fusion is positively related to both its fusion inputs (i.e., RGBSal and DSal), we attempt to further improve the performance of the RGB saliency subnet by connecting its encoder layers with the multi-scale deep features of the encoding layers in the D saliency subnet, and then we fine-tune this novel RGB saliency subnet as the Stage-3 in Fig.~\ref{fig:Training}.
Finally, we jointly fine-tune the entire network over the entire training set, including the MSF subnet as well (i.e., the stage 4 in Fig.~\ref{fig:Training}).

Specifically, the pseudo-GTs for the DCA subnet training may be ill-formulated if we use the entire training set to train the RGB and D saliency subnets.
To avoid such case, we equally divide the training set into 2 parts, one for the stage-wise training of the RGB and D saliency subnets, and another facilitate the formulation of the pseudo-GTs for the DCA subnet.

%{\rm (2)}&\ temp \gets \Big(1+Sigmoid\big(h(temp)\big)\Big)\times temp,\\
\section{Experiments}
\subsection{Datasets}
We have evaluated our approach on 5 publicly available datasets: NJUDS~\cite{ICIP_J2014}, NLPR~\cite{ECCV_P2014}, STEREO~\cite{SSB}, DES~\cite{DES}, LFSD~\cite{LFSD}.
The NJUDS~\cite{ICIP_J2014} dataset contains 1985 RGB-D images with well annotated binary saliency ground truth.
The NLPR~\cite{ECCV_P2014} dataset contains 1000 RGB-D images from either the indoor or outdoor scenes.
The STEREO~\cite{SSB} dataset includes 1000 images with low-quality D, DES~\cite{DES} dataset has 135 indoor pictures taken by Kinect, and LFSD~\cite{LFSD} dataset has 100 RGB-D images.

In order to make a fair comparison with the SOTA methods, we follow the same training/testing data split scheme as~\cite{zhao2019contrast}, in which we divide the NJUDS dataset (1985 images) into 2 parts, i.e., 1400 for training and the rest for testing; we also divide the NLPR dataset (1K images) into 2 parts, 650 for training and the rest for testing; all the STEREO, DES and LFSD datasets mentioned above will be used for testing.
Thus, our training set totally consists of 2050 images.

In addition, our RGB saliency subnet was pre-trained using two large-scale RGB datasets, i.e., the MSRA10K and DUTS datasets with a total of 20,553 RGB images; our D saliency subnet was pre-trained using the widely-used 2050 RGB-D images.
All training and testing images are resized to $224\times 224$.

\begin{figure*}[t]
\centering
\includegraphics[width=1\linewidth]{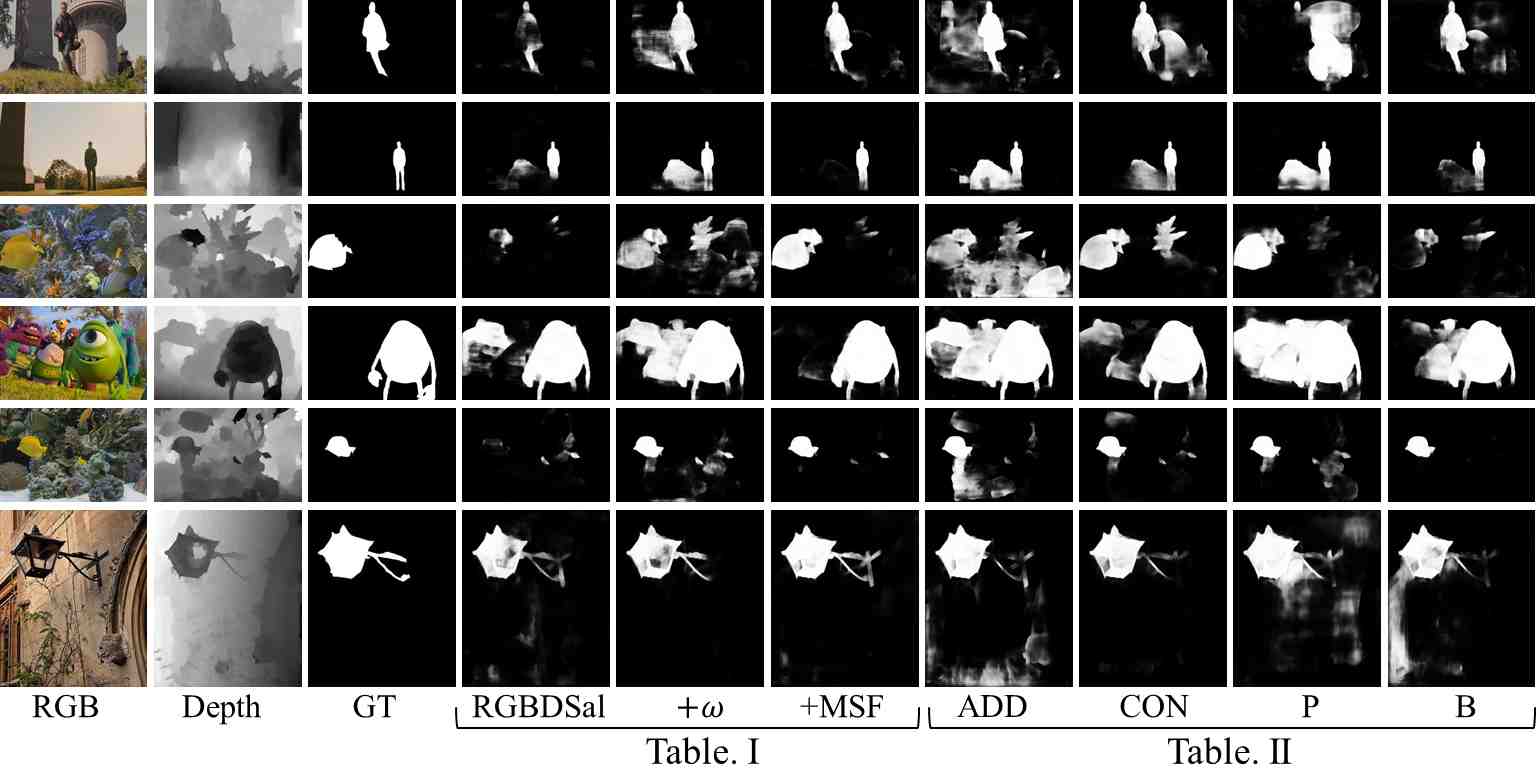}
\caption{The qualitative demonstrations of several important component mentioned in our ablation experiments.}
\label{fig:Ablation Experiments}
\end{figure*}

\subsection{Implementation Details}
Our experiments were performed on a workstation with a GTX1080Ti GPU (with 11G memory) and 64G RAM. Our network was implemented in Python 3.6 with TensorFlow1.4.
We set the batch size to 4 and adopt the ADAM optimizer to update and calculate network parameters.
All subnets in our method adopt the pre-trained VGG19 as the backbone net.
It takes about 7 hours (15 epochs) to pre-train our RGB saliency subnet over the above mentioned RGB training set (20K).
Then, it takes another 1 hour to fine-tune the D saliency subnet over the entire RGB-D training set (2K) using the D channel only (15 epochs).
Then, the DCA (Depth Contribution Assessment) subnet training takes almost 3 hours.
Finally, it takes about 1.6 hours (12 epochs with learning rate from 0.0001 to 0.00001) to jointly fine-tune the entire network.
Our method totally needs almost 0.08s for a single RGB-D image during the testing stage.

\subsection{Evaluation Metrics}
We adopt the standard metrics to conduct the quantitative evaluations, including S-measure~\cite{fan2017structure}, E-measure~\cite{Fan2018Enhanced}, F-measure and MAE.
The S-measure~\cite{fan2017structure} is a new structural similarity measure, which is defined as:
Eq.~\ref{eq:SMeasure}.
\begin{equation}
\label{eq:SMeasure}
\mathrm{S} =\alpha \times S_{o}+(1-\alpha) \times S_{r},
\end{equation}
where we set $\alpha=0.5$ to balance the region-aware (So) and object-aware (Sr) structural similarity.

%E-measure is also named Enhanced-alignment Measure~\cite{Fan2018Enhanced}. Unlike the pixel-level F-measure or the image-level S-measure, this novel measure combines the global image-level with local pixel-level information to make an improvement than the other meta-measures, and it can be formulated as:
%\begin{equation}
%\label{eq:EMeasure}
%\mathrm{E}=\frac{1}{w \times h} \sum_{x=1}^{w} \sum_{y=1}^{h} \phi_{\rm FM}(x, y),
%\end{equation}
%where $w$ and $h$ represent the width and height of the image respectively. FM represents the foreground map. $\phi$ is an
%enhanced alignment matrix to capture the pixel-level matching and image-level statistics.

The F-measure metric takes both precision and recall into account simultaneously, and its definition is shown as:
\begin{equation}
\label{eq:FMeasure}
\mathrm{F}=\frac{\left(\beta^{2}+1\right) \times \mathrm{Precision} \times \mathrm{Recall}}{\beta^{2} \times \mathrm{Precision}+\mathrm{Recall}},
\end{equation}
were we set ${\beta^{2}}$ = 0.3 as suggested in~\cite{achanta2009frequency}.

The MAE~\cite{zhang2017amulet} definition is shown as:
\begin{equation}
\label{eq:MAE}
\mathrm{MAE}=\frac{1}{T} \sum_{j=1}^{T}|{\rm S}_j -{\rm GT}_j|.
\end{equation}
where $\rm S$ and $\rm GT$ represent the saliency map and the saliency ground truth respectively.

\begin{table*}
\begin{center}
  %\centering
\caption{The overall component evaluation. $\uparrow$ denotes larger is better, and $\downarrow$ denotes smaller is better.}
  \resizebox{0.9\linewidth}{!}{
\smallskip\begin{tabular}{c|c|c|c|c|c|c|c|c|c|c|c|c|c|c|c}
\toprule[1pt]
    Dataset & \multicolumn{3}{c|}{NJDUS} & \multicolumn{3}{c|}{STEREO} & \multicolumn{3}{c|}{DES} & \multicolumn{3}{c|}{NLPR} & \multicolumn{3}{c}{LFSD} \\
    \hline
    Metric & Sm $\uparrow$ & meanF $\uparrow$ & MAE $\downarrow$   & Sm $\uparrow$    & meanF $\uparrow$ & MAE $\downarrow$   & Sm $\uparrow$    & meanF $\uparrow$ & MAE $\downarrow$   & Sm $\uparrow$    & meanF $\uparrow$ & MAE $\downarrow$   & Sm $\uparrow$    & meanF $\uparrow$ & MAE $\downarrow$ \\
    \midrule
    RGBSal & 0.856 & 0.801 & 0.080  & 0.883 & 0.825 & 0.063 & 0.874 & 0.792 & 0.047 & 0.896 & 0.819 & 0.042 & 0.808 & 0.756 & 0.120 \\
    DSal & 0.827 & 0.777 & 0.091 & 0.769 & 0.683 & 0.116 & 0.905 & 0.835 & 0.040  & 0.834 & 0.739 & 0.063 & 0.757 & 0.702 & 0.140 \\
    RGBDSal     & 0.885 & 0.850  & 0.056 & 0.883 & 0.838 & 0.056 & 0.920  & 0.870  & 0.028 & 0.908 & 0.853 & 0.033 & 0.827 & 0.796 & 0.102 \\
    $\rm RGBDSal^+$    & 0.874 & 0.840 & 0.061 & 0.870 & 0.823  & 0.060  & 0.911 & 0.873  & 0.027 & 0.901 & 0.851 & 0.033 & 0.789 & 0.76   & 0.110 \\
    +$\omega$   & 0.891 & 0.855 & 0.056 & 0.889 & 0.838 & 0.055 & 0.931 & 0.885 & 0.026 & 0.917 & 0.866 & 0.031 & 0.848 & 0.813 & 0.091 \\
    +MSF  & \textbf{0.897} & \textbf{0.873} & \textbf{0.052} & \textbf{0.892} & \textbf{0.854} & \textbf{0.051} & \textbf{0.935} & \textbf{0.901} & \textbf{0.021} & \textbf{0.916} & \textbf{0.864} & \textbf{0.029} & \textbf{0.851} & \textbf{0.826} & \textbf{0.085}\\
\toprule[1pt]
    %\hline
    \end{tabular}%
        \label{table:overallcomponent}
}
\end{center}
\vspace{-0.2cm}
\end{table*}

\begin{table*}
\begin{center}
  %\centering
\caption{Ablation study on the DCA subnet. $\uparrow$ denotes larger is better, and $\downarrow$ denotes smaller is better.}
  \resizebox{0.9\linewidth}{!}{
\smallskip\begin{tabular}{c|c|c|c|c|c|c|c|c|c|c|c|c|c|c|c}
\toprule[1pt]
    Dataset & \multicolumn{3}{c|}{NJDUS} & \multicolumn{3}{c|}{STEREO} & \multicolumn{3}{c|}{DES} & \multicolumn{3}{c|}{NLPR} & \multicolumn{3}{c}{LFSD} \\
    \hline
    Metric & Sm $\uparrow$ & meanF $\uparrow$ & MAE $\downarrow$   & Sm $\uparrow$    & meanF $\uparrow$ & MAE $\downarrow$   & Sm $\uparrow$    & meanF $\uparrow$ & MAE $\downarrow$   & Sm $\uparrow$    & meanF $\uparrow$ & MAE $\downarrow$   & Sm $\uparrow$    & meanF $\uparrow$ & MAE $\downarrow$ \\
\midrule
    ADD & 0.876 & 0.830  & 0.067 & 0.872 & 0.796 & 0.073 & 0.927 & 0.874 & 0.026 & 0.891 & 0.815 & 0.042 & 0.830  & 0.783 & 0.107 \\
    CON & 0.883 & 0.856 & 0.059 & 0.887 & 0.829 & 0.060  & 0.918 & 0.872 & 0.028 & 0.909 & 0.857 & 0.033 & 0.841 & 0.805 & 0.099 \\
    P (Eq.~\ref{eq:P}) & 0.889 & 0.854 & 0.056 & 0.888 & 0.841 & 0.054 & 0.927 & 0.883 & 0.027 & 0.911 & 0.855 & 0.033 & 0.844 & 0.808 & 0.093 \\
    B (Eq.~\ref{eq:B}) & 0.894 & 0.865 & 0.053 & 0.891 & 0.854 & 0.051 & 0.919 & 0.888 & 0.026 & 0.911 & 0.856 & 0.033 & 0.853 & 0.828 & 0.089 \\
    P+B  & \textbf{0.897} & \textbf{0.873} & \textbf{0.052} & \textbf{0.892} & \textbf{0.854} & \textbf{0.051} & \textbf{0.935} & \textbf{0.901} & \textbf{0.021} & \textbf{0.916} & \textbf{0.864} & \textbf{0.029} & \textbf{0.851} & \textbf{0.826} & \textbf{0.085} \\
\toprule[1pt]
    \end{tabular}%
        \label{table:DCA}
}
\end{center}
\vspace{-0.2cm}
\end{table*}

\begin{table*}
\begin{center}
  %\centering
\caption{Ablation study on the MSF subnet. $\uparrow$ denotes larger is better, and $\downarrow$ denotes smaller is better.}
\vspace{-0.2cm}
  \resizebox{1\linewidth}{!}{
\smallskip\begin{tabular}{c|c|c|c|c|c|c|c|c|c|c|c|c|c|c|c}
\toprule[1pt]
    Dataset & \multicolumn{3}{c|}{NJDUS} & \multicolumn{3}{c|}{STEREO} & \multicolumn{3}{c|}{DES} & \multicolumn{3}{c|}{NLPR} & \multicolumn{3}{c}{LFSD} \\
    \hline
    Metric & Sm $\uparrow$ & meanF $\uparrow$ & MAE $\downarrow$   & Sm $\uparrow$    & meanF $\uparrow$ & MAE $\downarrow$   & Sm $\uparrow$    & meanF $\uparrow$ & MAE $\downarrow$   & Sm $\uparrow$    & meanF $\uparrow$ & MAE $\downarrow$   & Sm $\uparrow$    & meanF $\uparrow$ & MAE $\downarrow$ \\
\midrule
    SimpleFusion & 0.877 & 0.839 & 0.062 & 0.875 & 0.821 & 0.061 & 0.916 & 0.863 & 0.032 & 0.898 & 0.834 & 0.037 & 0.833 & 0.795 & 0.103 \\
    $\omega$(RGB+D)     & 0.883 & 0.852 & 0.058 & 0.876 & 0.830  & 0.060  & 0.913 & 0.860  & 0.034 & 0.906 & 0.856 & 0.034 & 0.829 & 0.800   & 0.104 \\
    $\omega$(RGBD+D)   & 0.883 & 0.853 & 0.058 & 0.887 & 0.844 & 0.055 & 0.913 & 0.874 & 0.03  & 0.913 & 0.859 & 0.032 & 0.839 & 0.813 & 0.096 \\
    MSF(RGB+D) & 0.887 & 0.853 & 0.058 & 0.887 & 0.837 & 0.055 & 0.921 & 0.875 & 0.028 & 0.913 & 0.859 & 0.032 & 0.840  & 0.802 & 0.097 \\
    MSF(RGBD+D)  & \textbf{0.897} & \textbf{0.873} & \textbf{0.052} & \textbf{0.892} & \textbf{0.854} & \textbf{0.051} & \textbf{0.935} & \textbf{0.901} & \textbf{0.021} & \textbf{0.916} & \textbf{0.864} & \textbf{0.029} & \textbf{0.851} & \textbf{0.826} & \textbf{0.085} \\
\toprule[1pt]
    \end{tabular}%
        \label{table:MSF}
}
\end{center}
\end{table*}

\subsection{Component Evaluation}
\textbf{Overall component evaluation}.
To validate the effectiveness of each major component in our method, here, we have conducted an extensive component evaluation, including RGBSal, DSal, DCA (Depth Contribution Assessment) subnet (Sec.~\ref{sec:DQSN}), and the MSF (Multi-scale Fusion) subnet (Sec.~\ref{sec:MSDF}).
The detailed quantitative results over the adopted datasets can be found in Table~\ref{table:overallcomponent}, in which the ``RGBSal'' denotes the results of the RGB saliency subnet and the ``DSal'' denotes the results of the D saliency subnet.
Next, we have tested the conventional selective RGB-D fusion (denoted by the ``RGBDSal'' in Table~\ref{table:overallcomponent}) as~\cite{Zhu2018PDNet}, which effectively improves the overall detection performance significantly as expected.
Then, we further use the $\omega$ (predicted by the DCA subnet) to guide the feature-level fusion before conducting the selective deep fusion (i.e., Eq.~\ref{eq:FeatureFusion}), which outperforms the ``RGBDSal'' significantly, showing the effectiveness of our DCA subnet.

Specifically, the total network parameter size of the proposed ``triple-stream'' network is about 120M, while the conventional ``bi-stream'' RGBDSal network is about 88M.
To make our component evaluation more convincing, we have newly tested a new version ``RGBDSal'', in which we have increased the channel number of $M_{1-4}$ from 64 to 256, and thus the parameter size is increased from the original 88M to the current 120M, making this new network has almost the same parameter size as the proposed ``$+\omega$''.
We use ``$\rm RGBDSal^+$'' to denote the revised new version, whose performance has decreased slightly.
The main reason causing this result may be explained from the following aspect: as we all know, we may not always achieve a better performance by simply increasing network capacity, and, in sharp contrast, a heavy network design with more parameters may lead to difficulty in network training and even degenerate the overall performance.

Finally, we further conduct the feature-level fusion with multi-scale selective fusion network (``+MSF'') as mentioned in Sec.~\ref{sec:MSDF}, and it has achieved the best performance in all tested datasets, indicating the effectiveness of our novel multi-scale selective fusion network.

\begin{figure*}%%%%%%%%%%%%%%%%%%%%%%%%%%%%%%%%%%%%%%%%%%%%%%%%%%%%%%%%%%%%%%%%%%%result.png
\begin{center}
\includegraphics[width=1\linewidth]{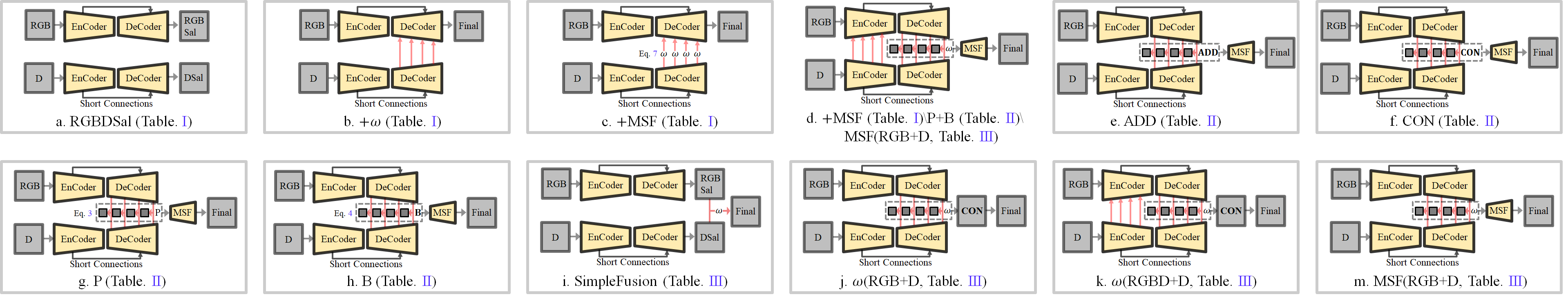}
\end{center}
   \caption{The detailed component evaluation network configurations, in which the ``Final'' denotes the final saliency detection results.}
\label{fig:ComFig}
\end{figure*}

\textbf{Ablation study on the DCA subnet}.
We have further conducted several experiments to show the effectiveness of our DCA (Depth Contribution Assessment) subnet (Sec.~\ref{sec:DQSN}), and the quantitative results can be found in Table~\ref{table:DCA}, in which ``$ADD$'' and ``$CON$'' respectively denote conducting the feature-level RGB-D fusion by using the addition operation and the $1\times 1$ convolutional operation, and the performance of these two schemes are limited due to their nature of depth quality-unaware;
``$P$'' and ``$B$'' respectively denote the results of training DCA subnet using either the P (Eq.~\ref{eq:P}) or B (Eq.~\ref{eq:B}) based pseudo-GT only, which have demonstrated slight performance improvement, but not much.
In fact, as for the salient object detection in RGB-D image, the main contribution of the D channel are two aspects: 1) the D channel helps the RGB channel to highlight the salient objects (i.e., Eq.~\ref{eq:P}); 2) more importantly, the D channel can effectively suppress those non-salient nearby surroundings (Eq.~\ref{eq:B}).
So, our DCA subnet may fail to measure the contribution of the D channel if we solely use either $P$ or $B$ during its training period.
We have demonstrated the final results, i.e., ``$P+B$'' in the bottom row of Table~\ref{table:DCA}, which shows the effectiveness of our DCA subnet.

\begin{table*}[t]
\begin{center}
%\centering
\caption{Quantitative evidence towards the effectiveness of the component B (Eq.~\ref{eq:B}).}
  \vspace{0.1cm}
\resizebox{0.9\linewidth}{!}{
\smallskip\begin{tabular}{c|c|c|c|c|c|c|c|c|c|c|c|c|c|c|c}
\toprule[1pt]
Dataset & \multicolumn{3}{c|}{NJDUS} & \multicolumn{3}{c|}{STEREO} & \multicolumn{3}{c|}{DES} & \multicolumn{3}{c|}{NLPR} & \multicolumn{3}{c}{LFSD} \\
\hline
w & P & B & P+B & P & B & P+B & P & B & P+B & P & B & P+B & P & B & P+B \\
\midrule
 $\omega_1$ & 29.9\% & 62.2\% & 45.9\% & 39.5\% & 66.5\% & 51.8\% & 29.9\% & 67.7\% & 50.0\% & 39.3\% & 69.7\% & 57.2\% & 30.1\% & 63.9\% & 45.1\% \\
 $\omega_2$ & 49.9\% & 74.5\% & 66.8\% & 30.9\% & 63.9\% & 52.3\% & 40.1\% & 71.7\% & 61.0\% & 50.2\% & 74.0\% & 65.5\% & 41.9\% & 71.3\% & 59.5\% \\
\toprule[1pt]
\end{tabular}%
            \label{table:B}
}
\end{center}
\end{table*}

\textbf{Ablation study on the MSF subnet}.
We have further conducted several experiments to show the effectiveness of our MSF (Multi-scale Fusion) subnet (Sec.~\ref{sec:MSDF}), and the quantitative results can be found in Table~\ref{table:MSF}.
Once the DCA subnet has been trained, we can directly use its prediction ($\omega$) to guide an explicit ``SimpleFusion'' between the stage-wise RGBSal and the stage-wise DSal as the Eq.~\ref{eq:SimpleFusion}.
Since such fusion scheme has only used the stage-wise RGBSal and DSal, it is difficult for such fusion to achieve the full complementary status between the RGB subnet and the D subnet (i.e., the multi-scale information), and thus this scheme has exhibited the worst performance in Table~\ref{table:MSF} ($SimpleFusion$).
Then, we have tested four different deep fusion schemes, in which these schemes are all based on the RGB-D deep features guided by the $\omega$ via Eq.~\ref{eq:FeatureFusion}.
The ``$\omega(RGB+D)$'' and the ``$\omega(RGBD+D)$'' denote two schemes which use the $\omega$ to obtain four fused side-outputs of its decoder layers first, and then these side-outputs will be simultaneously convolved as the final saliency output.
The major difference between the $\omega(RGB+D)$ and the $\omega(RGBD+D)$ is that the $\omega(RGB+D)$ do not use any encoder short-connections between its RGB subnet and D subnet, while the $\omega(RGBD+D)$ has connected its RGB subnet with D subnet as the gray lines in the left part of Fig.~\ref{fig:Net}.
The quantitative results ($\omega(RGBD+D)$$>$$\omega(RGB+D)$) have suggested that the complementary fusion between RGB and D can be benefited from both the following 2 aspects: 1) the classic RGB-D selective deep fusion at the encoder stage (to obtain RGBD saliency); 2) our $\omega$ guided selective deep fusion at the decoder layer (to complement the RGBD saliency with the D saliency).
Moreover, we have tested the performance of our MSF subnet, in which we feed the aforementioned four fused side-outputs into the MSF subnet to compute the final saliency map.
Similarly, we have tested two different schemes, i.e., ``$MSF(RGB+D)$'' and ``$MSF(RGBD+D)$'', which respectively correlate to the $\omega(RGB+D)$ and the $\omega(RGBD+D)$, showing the effectiveness of using RGBD saliency to replace RGB saliency during fusion.
Meanwhile, such results have also demonstrated the advantages of our MSF network than the conventional selective fusion scheme.

Specially, we have listed the detailed component evaluation network architecture in Fig.~\ref{fig:ComFig}.
Also, we have demonstrated the qualitative demonstrations of each key component in Fig.~\ref{fig:Ablation Experiments}.

\textbf{Effectiveness of B} (Eq.~\ref{eq:B}).
Compared with the ``RGBDSal'' that adopts the full automatic fusion scheme, the quantitative/qualitative results of ``$+\omega$'' are simply obtained by performing weighted summation between RGBSal and DSal---a typical hand-crafted manner, which has overlooked the ``multi-scale'' complementary nature between RGBSal and DSal, leading to limited performance frequently.
Nevertheless, because of its special advantage---depth quality aware, the ``$+\omega$'' can still outperform the RGBDSal in terms of both detection completeness and accuracy.

As shown in Fig.~\ref{fig:Ablation Experiments}, though the ``$+\omega$'' can improve the overall performance, it may occasionally produce more false-alarm detections in non-salient regions due to the following two reasons:\\
1) Since the ``$+\omega$'' has followed a hand-crafted methodology, the fusion procedure cannot take full advantage of the newly available depth contribution assessment map, i.e., the ``$\omega$''.\\
2) Moreover, the ``$\omega$'' itself still exists incorrect predictions (from both P and B), producing false-alarm detections easily in non-salient regions.\\
Thus, though P may bring more complete SOD, it easily introduces more false-alarm detections in non-salient regions, even in the case that B can compress these false-alarm detections in some extent.
In a word, the demonstrations in Fig.~\ref{fig:Ablation Experiments} are the results determined by both P and B jointly, and it is quite reasonable to illustrate more false-alarm detections occasionally.

Also, the ``B'' component is designed to predict those most valuable depth regions in non-salient regions, in which the value of depth channel in non-salient regions is just to suppress those false-alarm detections.
Therefore, the effectiveness of B towards suppressing non-salient regions only relies in the case whether those regions with strong feature responses in B must correlate to low feature responses in DSal.
To support this claim, we have provided an additional quantitative result towards the relationship between P, B and P+B in terms of suppressing non-salient image regions.

As shown in Table~\ref{table:B}, $\omega_1$ and $\omega_2$ are measured by Eq.~\ref{eq:omega1} and Eq.~\ref{eq:omega2} respectively.

\begin{equation}
\label{eq:omega1}
\omega_1 = \frac{||(1-GT)\odot(\omega-0.8)_+||_0}{||(\omega-0.8)_+||_0},
\end{equation}

\begin{equation}
\label{eq:omega2}
\omega_2 = \frac{||(1-GT)\odot(0.1-DSal)_+\odot(\omega-0.8)_+||_0}{||(1-GT)\odot(\omega-0.8)_+||_0},
\end{equation}
where $(\cdot)_+$ set all negative elements into zero; $||\cdot||_0$ denotes the $L_0$-norm; 0.1 and 0.8 are two predefined hard-thresholds.

The behind rationale of ``$\omega_1$'' is to measure the percentage of high-quality D regions predicted by the DCA in the non-salient regions.
The ``$\omega_2$'' further measures the percentage of regions in $\omega_1$ which can really compress non-salient regions.

From Table~\ref{table:B}, we can easily notice the following three aspects:\\
1) Compared with the component P, the component B tends to compress non-salient regions in general.\\
2) As expected, the combined P+B is better than P but worse than B in compressing the non-salient regions.\\
These two aspects can well explain why the ``$+\omega$'' (i.e., the P+B) may occasionally produce more false-alarm detections.

\begin{table*}[t]
  \centering
  \caption{Quantitative comparison results including S-measure, meanF and MAE on 5 public datasets. $\uparrow$ denotes larger is better, and $\downarrow$ denotes smaller is better. The top results are highlighted in bold font. NLR:~ NLPR~\cite{ECCV_P2014}; NJU:~NJUDS~\cite{ICIP_J2014}; O:~MSRA10K~\cite{liu2010learning} + DUTS\_TR~\cite{wang2017learning}.}
  \vspace{0.1cm}
  \resizebox{1\linewidth}{!}{
    \begin{tabular}{r|cc|ccc|ccc|ccc|ccc|ccc}
\toprule[1pt]
    Dataset & \multicolumn{2}{c|}{Training Details} & \multicolumn{3}{c|}{NJDUS} & \multicolumn{3}{c|}{STEREO} & \multicolumn{3}{c|}{DES} & \multicolumn{3}{c|}{NLPR} & \multicolumn{3}{c}{LFSD} \\
    \hline
    Metric & \multicolumn{1}{c}{Image Num.} & Dataset & \multicolumn{1}{c}{Sm $\uparrow$} & \multicolumn{1}{c}{meanF $\uparrow$} & MAE $\downarrow$  & \multicolumn{1}{c}{Sm $\uparrow$} & \multicolumn{1}{c}{meanF $\uparrow$} & MAE $\downarrow$  & \multicolumn{1}{c}{Sm $\uparrow$} & \multicolumn{1}{c}{meanF $\uparrow$} & MAE $\downarrow$  & \multicolumn{1}{c}{Sm $\uparrow$} & \multicolumn{1}{c}{meanF $\uparrow$} & MAE $\downarrow$  & \multicolumn{1}{c}{Sm $\uparrow$} & \multicolumn{1}{c}{meanF $\uparrow$} & MAE $\downarrow$ \\
    \midrule
    CPD19~\cite{CPD} & \multicolumn{1}{c}{0.70K+1.5K} & NLR+NJU & \multicolumn{1}{c}{0.872} & \multicolumn{1}{c}{0.847} & 0.059 & \multicolumn{1}{c}{0.888} & \multicolumn{1}{c}{\textbf{0.856}} & 0.050 & \multicolumn{1}{c}{0.877} & \multicolumn{1}{c}{0.849} & 0.038 & \multicolumn{1}{c}{0.901} & \multicolumn{1}{c}{0.860} & 0.034 & \multicolumn{1}{c}{0.797} & 0.771 & 0.112 \\
    EGNet19~\cite{EGNet} & \multicolumn{1}{c}{0.70K+1.5K} & NLR+NJU & \multicolumn{1}{c}{0.818} & \multicolumn{1}{c}{0.784} & 0.088 & \multicolumn{1}{c}{0.826} & \multicolumn{1}{c}{0.786} & 0.079 & \multicolumn{1}{c}{0.811} & \multicolumn{1}{c}{0.777} & 0.058 & \multicolumn{1}{c}{0.850} & \multicolumn{1}{c}{0.796} & 0.051 & \multicolumn{1}{c}{0.816} & 0.790 & 0.103 \\
    BASNet19~\cite{BASNet} & \multicolumn{1}{c}{0.70K+1.5K} & NLR+NJU & \multicolumn{1}{c}{0.826} & \multicolumn{1}{c}{0.797} & 0.082 & \multicolumn{1}{c}{0.857} & \multicolumn{1}{c}{0.823} & 0.063 & \multicolumn{1}{c}{0.826} & \multicolumn{1}{c}{0.772} & 0.052 & \multicolumn{1}{c}{0.889} & \multicolumn{1}{c}{0.841} & 0.037 & \multicolumn{1}{c}{0.748} & 0.717 & 0.132 \\
    \midrule
    MDSF17~\cite{song2017depth} & \multicolumn{1}{c}{0.50K+0.5K} & NLR+NJU & \multicolumn{1}{c}{0.748} & \multicolumn{1}{c}{0.628} & 0.157 & \multicolumn{1}{c}{0.728} & \multicolumn{1}{c}{0.527} & 0.176 & \multicolumn{1}{c}{0.741} & \multicolumn{1}{c}{0.523} & 0.122 & \multicolumn{1}{c}{0.805} & \multicolumn{1}{c}{0.649} & 0.095 & \multicolumn{1}{c}{0.700} & 0.521 & 0.190 \\
    DF17~\cite{TIP_Q2017}  & \multicolumn{1}{c}{0.75K+1.0K} & NLR+NJU & \multicolumn{1}{c}{0.763} & \multicolumn{1}{c}{0.650} & 0.141 & \multicolumn{1}{c}{0.757} & \multicolumn{1}{c}{0.617} & 0.141 & \multicolumn{1}{c}{0.752} & \multicolumn{1}{c}{0.604} & 0.093 & \multicolumn{1}{c}{0.802} & \multicolumn{1}{c}{0.664} & 0.085 & \multicolumn{1}{c}{0.791} & 0.679 & 0.138 \\
    CTMF18~\cite{han2017cnns} & \multicolumn{1}{c}{0.65K+1.4K} & NLR+NJU & \multicolumn{1}{c}{0.849} & \multicolumn{1}{c}{0.779} & 0.085 & \multicolumn{1}{c}{0.848} & \multicolumn{1}{c}{0.758} & 0.086 & \multicolumn{1}{c}{0.863} & \multicolumn{1}{c}{0.756} & 0.055 & \multicolumn{1}{c}{0.860} & \multicolumn{1}{c}{0.740} & 0.056 & \multicolumn{1}{c}{0.796} & 0.756 & 0.119 \\
    PCF18~\cite{chen2018progressively} & \multicolumn{1}{c}{0.70K+1.5K} & NLR+NJU & \multicolumn{1}{c}{0.877} & \multicolumn{1}{c}{0.840} & 0.059 & \multicolumn{1}{c}{0.875} & \multicolumn{1}{c}{0.818} & 0.064 & \multicolumn{1}{c}{0.842} & \multicolumn{1}{c}{0.765} & 0.049 & \multicolumn{1}{c}{0.874} & \multicolumn{1}{c}{0.802} & 0.044 & \multicolumn{1}{c}{0.794} & 0.761 & 0.112 \\
    PDNet18~\cite{Zhu2018PDNet} & \multicolumn{1}{c}{0.70K+1.5K+21K} & NLR+NJU+O & \multicolumn{1}{c}{0.877} & \multicolumn{1}{c}{0.814} & 0.071 & \multicolumn{1}{c}{0.830} & \multicolumn{1}{c}{0.730} & 0.092 & \multicolumn{1}{c}{0.887} & \multicolumn{1}{c}{0.795} & 0.045 & \multicolumn{1}{c}{0.887} & \multicolumn{1}{c}{0.802} & 0.050 & \multicolumn{1}{c}{0.847} & 0.779 & 0.107 \\
    AFNet19~\cite{wang2019adaptive} & \multicolumn{1}{c}{0.70K+1.5K} & NLR+NJU & \multicolumn{1}{c}{0.772} & \multicolumn{1}{c}{0.764} & 0.100 & \multicolumn{1}{c}{0.825} & \multicolumn{1}{c}{0.806} & 0.075 & \multicolumn{1}{c}{0.770} & \multicolumn{1}{c}{0.713} & 0.068 & \multicolumn{1}{c}{0.799} & \multicolumn{1}{c}{0.755} & 0.058 & \multicolumn{1}{c}{0.738} & 0.735 & 0.133 \\
    MMCI19~\cite{chen2019multi} & \multicolumn{1}{c}{0.70K+1.5K} & NLR+NJU & \multicolumn{1}{c}{0.858} & \multicolumn{1}{c}{0.793} & 0.079 & \multicolumn{1}{c}{0.873} & \multicolumn{1}{c}{0.813} & 0.068 & \multicolumn{1}{c}{0.848} & \multicolumn{1}{c}{0.735} & 0.065 & \multicolumn{1}{c}{0.856} & \multicolumn{1}{c}{0.737} & 0.059 & \multicolumn{1}{c}{0.787} & 0.722 & 0.132 \\
    TANet19~\cite{chen2019three} & \multicolumn{1}{c}{0.70K+1.5K} & NLR+NJU & \multicolumn{1}{c}{0.878} & \multicolumn{1}{c}{0.841} & 0.060 & \multicolumn{1}{c}{0.871} & \multicolumn{1}{c}{0.828} & 0.060 & \multicolumn{1}{c}{0.858} & \multicolumn{1}{c}{0.790} & 0.046 & \multicolumn{1}{c}{0.886} & \multicolumn{1}{c}{0.819} & 0.041 & \multicolumn{1}{c}{0.801} & 0.771 & 0.111 \\
    CPFP19~\cite{zhao2019contrast} & \multicolumn{1}{c}{0.70K+1.5K} & NLR+NJU & \multicolumn{1}{c}{0.878} & \multicolumn{1}{c}{0.850} & 0.053 & \multicolumn{1}{c}{0.879} & \multicolumn{1}{c}{0.841} & 0.051 & \multicolumn{1}{c}{0.872} & \multicolumn{1}{c}{0.824} & 0.038 & \multicolumn{1}{c}{0.888} & \multicolumn{1}{c}{0.840} & 0.036 & \multicolumn{1}{c}{0.828} & 0.811 & 0.088 \\
    \rowcolor{mygray}
    Ours(ResNet) & {0.70K+1.5K} & NLR+NJU & {0.889} & {0.863} & \textbf{0.051} & {0.880} & {0.846} & \textbf{0.049} & {0.912} & {0.884} & 0.025 & {0.903} & {\textbf{0.866}} & 0.032 & {0.831} & 0.810 & 0.086 \\
    Ours(VGG) & \multicolumn{1}{c}{0.70K+1.5K+21K} & NLR+NJU+O & \multicolumn{1}{c}{\textbf{0.897}} & \multicolumn{1}{c}{\textbf{0.873}} & 0.052 & \multicolumn{1}{c}{\textbf{0.892}} & \multicolumn{1}{c}{0.854} & 0.051 & \multicolumn{1}{c}{\textbf{0.935}} & \multicolumn{1}{c}{\textbf{0.901}} & \textbf{0.021} & \multicolumn{1}{c}{\textbf{0.916}} & \multicolumn{1}{c}{0.864} & \textbf{0.029} & \multicolumn{1}{c}{\textbf{0.851}} & \textbf{0.826} & \textbf{0.085} \\
\toprule[1pt]
    \end{tabular}%
}
\label{table:All}
\end{table*}%

\begin{figure*}%%%%%%%%%%%%%%%%%%%%%%%%%%%%%%%%%%%%%%%%%%%%%%%%%%%%%%%%%%%%%%%%%%%result.png
\begin{center}
\includegraphics[width=1\linewidth]{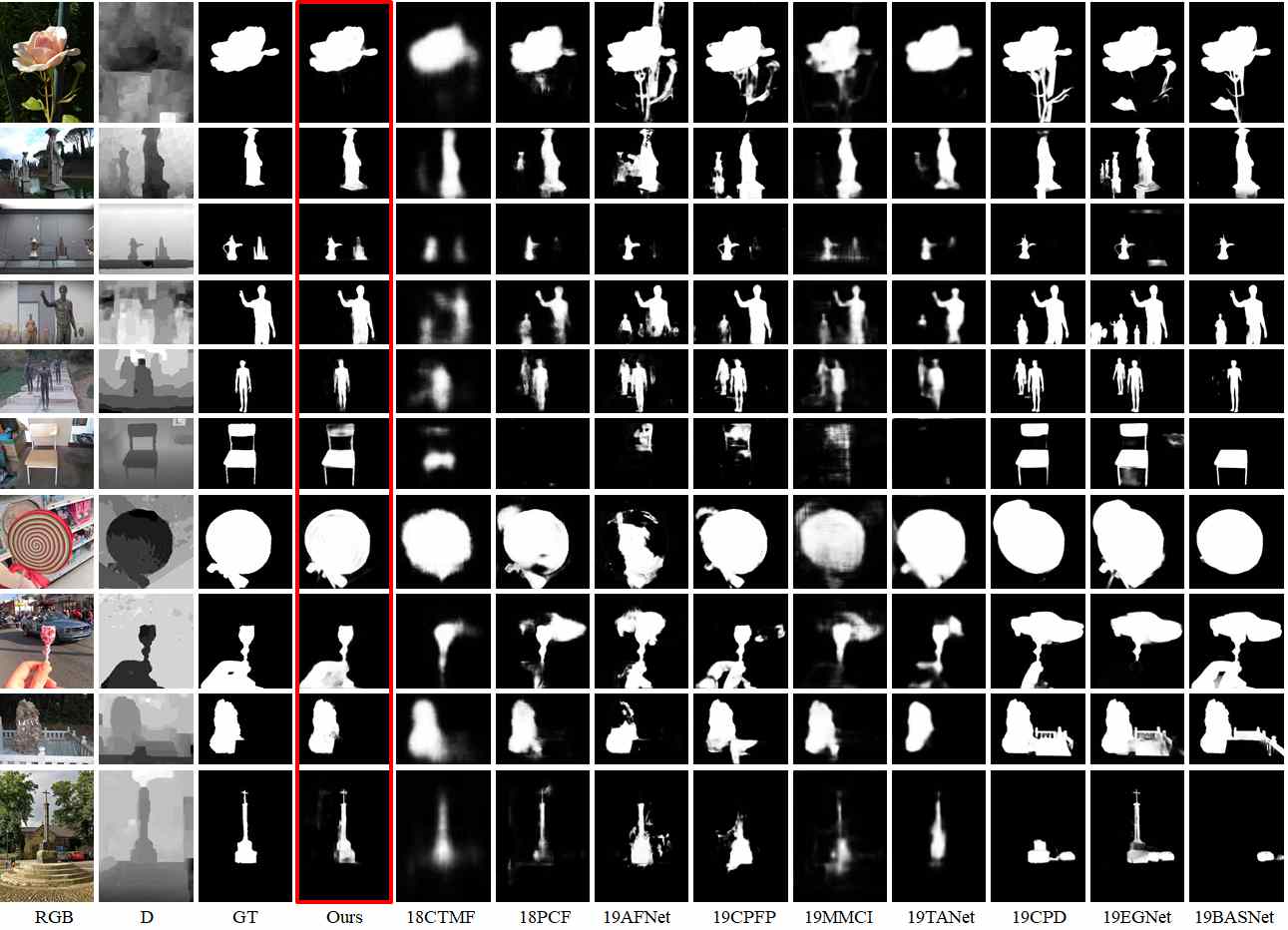}
\end{center}
   \caption{Qualitative comparisons with the SOTA methods.}
\label{fig:result}
\end{figure*}

\begin{table*}[htbp]
  \centering
  \caption{Quantitative results using different feature backbones. The top three are highlighted in \textbf{\color{red}red}, {\color{blue}blue} and {\color{green}green}. ``*'' denotes that the model is initially pre-trained using additional 20K RGB images.}
  \vspace{0.1cm}
  \resizebox{0.9\linewidth}{!}{
    \begin{tabular}{c|ccc|ccc|ccc|ccc|ccc}
\toprule[1pt]
    \multirow{2}[4]{*}{Method} & \multicolumn{3}{c|}{NJUDS} & \multicolumn{3}{c|}{STEREO} & \multicolumn{3}{c|}{DES} & \multicolumn{3}{c|}{NLPR} & \multicolumn{3}{c}{LFSD} \\
\cmidrule{2-16}          & \multicolumn{1}{c}{Sm $\uparrow$} & \multicolumn{1}{c}{meanF $\uparrow$} & MAE $\downarrow$   & \multicolumn{1}{c}{Sm $\uparrow$} & \multicolumn{1}{c}{meanF $\uparrow$} & MAE $\downarrow$   & \multicolumn{1}{c}{Sm $\uparrow$} & \multicolumn{1}{c}{meanF $\uparrow$} & MAE  $\downarrow$  & \multicolumn{1}{c}{Sm $\uparrow$} & \multicolumn{1}{c}{meanF $\uparrow$} & MAE $\downarrow$   & Sm $\uparrow$   & meanF $\uparrow$ & MAE $\downarrow$  \\
    \midrule
    CPFP19 & \multicolumn{1}{c}{0.878} & \multicolumn{1}{c}{0.850} & \textcolor[rgb]{0.000, 1.000, 0.000}{0.053} & \multicolumn{1}{c}{0.879} & \multicolumn{1}{c}{0.841} & 0.051 & \multicolumn{1}{c}{0.872} & \multicolumn{1}{c}{0.824} & 0.038 & \multicolumn{1}{c}{0.888} & \multicolumn{1}{c}{0.840} & 0.036 & 0.828 & \textcolor[rgb]{0.000, 0.000, 1.000}{0.811} & \textcolor[rgb]{0.000, 1.000, 0.000}{0.088} \\
    Ours(VGG) & \multicolumn{1}{c}{0.854} & \multicolumn{1}{c}{0.814} & 0.074 & \multicolumn{1}{c}{0.855} & \multicolumn{1}{c}{0.808} & 0.068 & \multicolumn{1}{c}{0.915} & \multicolumn{1}{c}{0.884} & 0.027 & \multicolumn{1}{c}{0.875} & \multicolumn{1}{c}{0.817} & 0.046 & 0.796 & 0.768 & 0.112 \\
    Ours(Res2Net) & \multicolumn{1}{c}{0.883} & \multicolumn{1}{c}{0.857} & 0.054 & \multicolumn{1}{c}{0.877} & \multicolumn{1}{c}{0.840} & 0.053 & \multicolumn{1}{c}{\textcolor[rgb]{0.000, 1.000, 0.000}{0.917}} & \multicolumn{1}{c}{0.886} & \textcolor[rgb]{0.000, 1.000, 0.000}{0.025} & \multicolumn{1}{c}{0.901} & \multicolumn{1}{c}{0.860} & 0.034 & 0.827 & 0.808 & 0.095 \\
    Ours*(ResNet) & \multicolumn{1}{c}{\textcolor[rgb]{0.000, 1.000, 0.000}{0.886}} & \multicolumn{1}{c}{\textcolor[rgb]{0.000, 0.000, 1.000}{0.865}} & 0.057 & \multicolumn{1}{c}{\textcolor[rgb]{0.000, 1.000, 0.000}{0.891}} & \multicolumn{1}{c}{\textcolor[rgb]{0.000, 0.000, 1.000}{0.860}} & \textcolor[rgb]{0.000, 0.000, 1.000}{0.047} & \multicolumn{1}{c}{0.913} & \multicolumn{1}{c}{\textcolor[rgb]{0.000, 1.000, 0.000}{0.888}} & 0.025 & \multicolumn{1}{c}{\textcolor[rgb]{0.000, 1.000, 0.000}{0.910}} & \multicolumn{1}{c}{\textcolor[rgb]{0.000, 0.000, 1.000}{0.868}} & \textcolor[rgb]{0.000, 1.000, 0.000}{0.030} & \textcolor[rgb]{0.000, 1.000, 0.000}{0.829} & 0.810 & 0.093 \\
    Ours*(Res2Net) & \multicolumn{1}{c}{0.883} & \multicolumn{1}{c}{\textcolor[rgb]{0.000, 1.000, 0.000}{0.864}} & 0.055 & \multicolumn{1}{c}{\textcolor[rgb]{0.000, 0.000, 1.000}{0.891}} & \multicolumn{1}{c}{\textcolor[rgb]{1.000, 0.000, 0.000}{0.863}} & \textcolor[rgb]{1.000, 0.000, 0.000}{0.046} & \multicolumn{1}{c}{\textcolor[rgb]{0.000, 0.000, 1.000}{0.933}} & \multicolumn{1}{c}{\textcolor[rgb]{1.000, 0.000, 0.000}{0.910}} & \textcolor[rgb]{1.000, 0.000, 0.000}{0.020} & \multicolumn{1}{c}{\textcolor[rgb]{0.000, 0.000, 1.000}{0.912}} & \multicolumn{1}{c}{\textcolor[rgb]{1.000, 0.000, 0.000}{0.874}} & \textcolor[rgb]{0.000, 0.000, 1.000}{0.030} & 0.824 & 0.806 & 0.094 \\
    \rowcolor{mygray}
    Ours(ResNet) & \textcolor[rgb]{0.000, 0.000, 1.000}{0.889} & 0.863 & \textcolor[rgb]{1.000, 0.000, 0.000}{0.051} & 0.880 & 0.846 & \textcolor[rgb]{0.000, 1.000, 0.000}{0.049} & 0.912 & 0.884 & 0.025 & 0.903 & \textcolor[rgb]{0.000, 1.000, 0.000}{0.866} & 0.032 & \textcolor[rgb]{0.000, 0.000, 1.000}{0.831} & \textcolor[rgb]{0.000, 1.000, 0.000}{0.810} & \textcolor[rgb]{0.000, 0.000, 1.000}{0.086} \\
    Ours*(VGG) & \multicolumn{1}{c}{\textcolor[rgb]{1.000, 0.000, 0.000}{0.897}} & \multicolumn{1}{c}{\textcolor[rgb]{1.000, 0.000, 0.000}{0.873}} & \textcolor[rgb]{0.000, 0.000, 1.000}{0.052} & \multicolumn{1}{c}{\textcolor[rgb]{1.000, 0.000, 0.000}{0.892}} & \multicolumn{1}{c}{\textcolor[rgb]{0.000, 1.000, 0.000}{0.854}} & 0.051 & \multicolumn{1}{c}{\textcolor[rgb]{1.000, 0.000, 0.000}{0.935}} & \multicolumn{1}{c}{\textcolor[rgb]{0.000, 0.000, 1.000}{0.901}} & \textcolor[rgb]{0.000, 0.000, 1.000}{0.021} & \multicolumn{1}{c}{\textcolor[rgb]{1.000, 0.000, 0.000}{0.916}} & \multicolumn{1}{c}{0.864} & \textcolor[rgb]{1.000, 0.000, 0.000}{0.029} & \textcolor[rgb]{1.000, 0.000, 0.000}{0.851} & \textcolor[rgb]{1.000, 0.000, 0.000}{0.826} & \textcolor[rgb]{1.000, 0.000, 0.000}{0.085} \\
\toprule[1pt]
    \end{tabular}%
}
  \label{tab:addlabel}%
\end{table*}%

\subsection{Comparison with the SOTA Methods}
We have compared our method with 12 most recent SOTA (3 top-level RGB methods and 9 RGB-D methods) methods over the aforementioned 5 datasets: 1) the compared RGB methods include CPD19~\cite{CPD}, EGNet19~\cite{EGNet} and BASNet19~\cite{BASNet}; 2) the compared RGB-D methods include MDSF17~\cite{song2017depth}, DF17~\cite{TIP_Q2017}, PDNet18~\cite{Zhu2018PDNet}, CTMF18~\cite{han2017cnns}, PCF18~\cite{chen2018progressively}, AFNet19~\cite{wang2019adaptive}, MMCI19~\cite{chen2019multi}, TANet19~\cite{chen2019three}, and CPFP19~\cite{zhao2019contrast}.

We have demonstrated the detailed quantitative comparison results in Table~\ref{table:All}, and the qualitative comparisons can be found in Fig.~\ref{fig:result}.
As shown in Table~\ref{table:All}, our method has achieved the best performance in all the tested datasets except the STEREO dataset.
Since the depth maps of STEREO are frequently with extremely low-quality, which cannot separate the salient objects from their non-salient surroundings nearby in most case. Thus, the performance gain came from the DSal subbranch, which meant to complement the RGBSal for a better overall performance, may get vanished and lead to inferior performance than the conventional color saliency models such as BASNet19.
Moreover, in sharp contrast to the conventional SOTA RGB-D salient object detection methods, our method is depth quality aware, which is able to bias the fusion balance to its RGB subnet in the STEREO dataset, avoiding the performance degeneration.

\subsection{Different Feature Backbones}
We have newly tested our method by using other two feature extractors, i.e., ResNet~\cite{he2016deep} and Res2Net~\cite{gao2019res2net}.
As shown in Table~\ref{tab:addlabel}, we have listed 7 (6 Ours + 1 CPFP, where we use the CPFP as the reference) quantitative results after using different feature backbones and different training strategies, including \{VGG, VGG*\}, \{ResNet50, ResNet50*\}, and \{Res2Net, Res2Net*\}, where ``*'' denotes that the model is initially pre-trained using additional 20K RGB images.

We have noticed that the pre-training process using 20K RGB images will improve the performance of the VGG based model significantly, while such improvements become marginal towards the ResNet and Res2Net based models.

For a fair comparison, we choose the ``Ours(ResNet)'' to represent the overall performance of the proposed method, which is trained without using any additional RGB images.
Compared with the vanilla CPFP19, our method outperforms it in all tested datasets.
Specifically, our method outperforms the CPFP19 in the DES dataset significantly.

\subsection{Limitations}
Compared with the conventional selective bi-stream fusion methods (e.g., PDNet), the key innovation of the proposed method is to devise a novel depth quality aware venue to eliminate side-effects from the low-quality depth information by biasing more towards the DSal during the fusion process.
Thus, the overall performance of our method is heavily dependent on its subbranches, i.e., the RGBSal subbranch and the DSal subbranch.
As a result, our method cannot perform well when both RGB and D streams could not well detect the salient objects, which is clearly a common problem of the bi-stream based SOD methods.

\section{Conclusion}
This paper has proposed a novel RGB-D salient object detection methods.
We have followed the widely used bi-stream structure as the baseline network, however, the bi-stream structure has one major limitation: it is unaware of the depth quality, which may limit its performance in facing of images with low-quality D.
To handle this limitation, we have adopted the weakly supervised learning scheme to train a novel subnet named DCA (Depth Contribution Assessment).
This novel DCA subnet is able to guide an explicit feature-level RGB-D fusion before conducting the selective deep fusion, making the conventional RGB-D bi-stream structure to become depth quality aware.
Moreover, we have introduced a novel selective deep fusion scheme to take full advantage of the DCA based multi-scale complementary information between the RGB subnet and the D subnet, achieving a much improved RGB-D salient object detection.
Finally, we have conducted massive quantitative evaluation to validate the effectiveness of our method.

\textbf{Acknowledgments}. This research is supported in part by
National Natural Science Foundation of China (No. 61802215
and No. 61806106), Natural Science Foundation of Shandong
Province (No. ZR2019BF011 and ZR2019QF009) and National Science Foundation of USA (No. IIS-1715985 and IIS-
1812606)
%\section{Conclusion}
%This paper has proposed a novel weakly supervised scheme to adapt image saliency deep models for video data. Our method can generate a novel motion saliency sub-branch via fine-tuning the off-the-shelf image saliency deep model using the color-coded optical flow data.
%Meanwhile, we propose the newly-designed key frame strategy to locate those frames with high-quality spatiotemporal saliency predictions.
%Then, we have used these high-quality predictions as the pseudo ground truth for the weakly supervised online training, which enables any off-the-shelf image saliency deep models to adapt for the current video sequence as the new color sub-branch of our method.
%Our method is simple, flexible, and effective, which is potentially to inspire future work even in the case that our color model adapted method is just comparable to the current leading state-of-the-art video saliency detection methods.

%\vspace{-0.4cm}

%\vspace{0.4cm}
%\textbf{Acknowledgments}. This research is supported in part by National Key R\&D Program of China (No. 2017YFF0106407), National Natural Science Foundation of China (No. 61802215 and No. 61806106), Natural Science Foundation of Shandong Province (No. ZR201807120086) and National Science Foundation of USA (No. IIS-1715985, IIS0949467, IIS-1047715, and IIS-1049448).

%\ifCLASSOPTIONcaptionsoff
%  \newpage
%\fi
%\vspace{0.4cm}
\bibliographystyle{IEEEtran}
\bibliography{reference}

\begin{thebibliography}{10}
\providecommand{\url}[1]{#1}
\csname url@samestyle\endcsname
\providecommand{\newblock}{\relax}
\providecommand{\bibinfo}[2]{#2}
\providecommand{\BIBentrySTDinterwordspacing}{\spaceskip=0pt\relax}
\providecommand{\BIBentryALTinterwordstretchfactor}{4}
\providecommand{\BIBentryALTinterwordspacing}{\spaceskip=\fontdimen2\font plus
\BIBentryALTinterwordstretchfactor\fontdimen3\font minus
  \fontdimen4\font\relax}
\providecommand{\BIBforeignlanguage}[2]{{%
\expandafter\ifx\csname l@#1\endcsname\relax
\typeout{** WARNING: IEEEtran.bst: No hyphenation pattern has been}%
\typeout{** loaded for the language `#1'. Using the pattern for}%
\typeout{** the default language instead.}%
\else
\language=\csname l@#1\endcsname
\fi
#2}}
\providecommand{\BIBdecl}{\relax}
\BIBdecl

\bibitem{fan2018salient}
D.~Fan, M.~Cheng, J.~Liu, S.~Gao, Q.~Hou, and A.~Borji, ``Salient objects in
  clutter: Bringing salient object detection to the foreground,'' in
  \emph{Proc. IEEE Eur. Conf. Comput. Vis. (ECCV)}, 2018, pp. 186--202.

\bibitem{CC2019TIP}
C.~Chen, G.~Wang, C.~Peng, X.~Zhang, and H.~Qin, ``Improved robust video
  saliency detection based on long-term spatial-temporal information,''
  \emph{IEEE Trans. on Image Process. (TIP)}, vol.~29, pp. 1090--1100, 2019.

\bibitem{CC2017TIP}
C.~Chen, S.~Li, Y.~Wang, H.~Qin, and A.~Hao, ``Video saliency detection via
  spatial-temporal fusion and low-rank coherency diffusion,'' \emph{IEEE Trans.
  on Image Process. (TIP)}, vol.~26, no.~7, pp. 3156--3170, 2017.

\bibitem{CC2019TMM2}
Y.~Li, S.~Li, C.~Chen, H.~Qin, and A.~Hao, ``Accurate and robust video saliency
  detection via selfpaced diffusion,'' \emph{IEEE Trans. on Multimedia (TMM)},
  p. early access, 2019.

\bibitem{CC2018TMM}
C.~Chen, S.~Li, H.~Qin, Z.~Pan, and G.~Yang, ``Bi-level feature learning for
  video saliency detection,'' \emph{IEEE Trans. on Multimedia (TMM)}, vol.~20,
  no.~12, pp. 3324--3336, 2018.

\bibitem{fan2019shifting}
D.~Fan, W.~Wang, M.~Cheng, and J.~Shen, ``Shifting more attention to video
  salient object detection,'' in \emph{Proc. IEEE Conf. Comput. Vis. Pattern
  Recognit. (CVPR)}, 2019, pp. 8554--8564.

\bibitem{zhang2015application}
W.~Zhang, A.~Borji, Z.~Wang, P.~Le~Callet, and H.~Liu, ``The application of
  visual saliency models in objective image quality assessment: A statistical
  evaluation,'' \emph{IEEE Trans. on Neural Netw. Learn. Syst. (TNNLS)},
  vol.~27, no.~6, pp. 1266--1278, 2015.

\bibitem{CC2015PR}
C.~Chen, S.~Li, H.~Qin, and A.~Hao, ``Real-time and robust object tracking in
  video via low-rank coherency analysis in feature space,'' \emph{Pattern
  Recognit. (PR)}, vol.~48, pp. 2885--2905, 2015.

\bibitem{CC2016PR}
C.~Chen, S.~Li, A.~Hao, and H.~Qin, ``Robust salient motion detection in
  non-stationary videos via novel integrated strategies of spatio-temporal
  coherency clues and low-rank analysis,'' \emph{Pattern Recognit. (PR)},
  vol.~52, pp. 410--432, 2016.

\bibitem{CC2019CVPR}
C.~Peng, C.~Chen, Z.~Kang, J.~Li, and Q.~Cheng, ``Res-pca: A scalable approach
  to recovering low-rank matrices,'' in \emph{Proc. IEEE Conf. Comput. Vis.
  Pattern Recognit. (CVPR)}, 2019.

\bibitem{CC2019TMM1}
G.~Ma, C.~Chen, S.~Li, C.~Peng, A.~Hao, and H.~Qin, ``Salient object detection
  via multiple instance joint re-learning,'' \emph{IEEE Trans. on Multimedia
  (TMM)}, 2019.

\bibitem{CC2015TIP}
C.~Chen, S.~Li, Y.~Wang, H.~Qin, and A.~Hao, ``Structure-sensitive saliency
  detection via multilevel rank analysis in intrinsic feature space,''
  \emph{IEEE Trans. on Image Process. (TIP)}, vol.~24, no.~8, pp. 2303--2316,
  2015.

\bibitem{ren2015exploiting}
J.~Ren, X.~Gong, L.~Yu, W.~Zhou, and M.~Ying~Yang, ``Exploiting global priors
  for rgb-d saliency detection,'' in \emph{Proc. IEEE Conf. Comput. Vis.
  Pattern Recognit. (CVPR)}, 2015, pp. 25--32.

\bibitem{chen2018progressively}
H.~Chen and Y.~Li, ``Progressively complementarity-aware fusion network for
  rgb-d salient object detection,'' in \emph{Proc. IEEE Conf. Comput. Vis.
  Pattern Recognit. (CVPR)}, 2018, pp. 3051--3060.

\bibitem{wang2019adaptive}
N.~Wang and X.~Gong, ``Adaptive fusion for rgb-d salient object detection,''
  \emph{IEEE Access}, vol.~7, pp. 55\,277--55\,284, 2019.

\bibitem{han2018cnns-based}
J.~Han, H.~Chen, N.~Liu, C.~Yan, and X.~Li, ``Cnns-based rgb-d saliency
  detection via cross-view transfer and multiview fusion,'' \emph{IEEE
  Transactions on Systems, Man, and Cybernetics}, vol.~48, no.~11, pp.
  3171--3183, 2018.

\bibitem{shigematsu2017learning}
R.~Shigematsu, D.~Feng, S.~You, and N.~Barnes, ``Learning rgb-d salient object
  detection using background enclosure, depth contrast, and top-down
  features,'' in \emph{Proc. IEEE Int. Conf. Comput. Vis. W. (ICCVW)}, 2017,
  pp. 2749--2757.

\bibitem{han2017cnns}
J.~Han, H.~Chen, N.~Liu, C.~Yan, and X.~Li, ``Cnns-based rgb-d saliency
  detection via cross-view transfer and multiview fusion,'' \emph{IEEE Trans.
  on Cybern. (TCYB)}, vol.~48, no.~11, pp. 3171--3183, 2018.

\bibitem{jiang2013salient}
H.~Jiang, J.~Wang, Z.~Yuan, Y.~Wu, N.~Zheng, and S.~Li, ``Salient object
  detection: A discriminative regional feature integration approach,'' in
  \emph{Proc. IEEE Conf. Comput. Vis. Pattern Recognit. (CVPR)}, 2013, pp.
  2083--2090.

\bibitem{peng2016salient}
H.~Peng, B.~Li, H.~Ling, W.~Hu, W.~Xiong, and S.~J. Maybank, ``Salient object
  detection via structured matrix decomposition,'' \emph{IEEE Trans. on Pattern
  Anal. Mach. Intell. (TPAMI)}, vol.~39, no.~4, pp. 818--832, 2016.

\bibitem{cong2017co}
R.~Cong, J.~Lei, H.~Fu, Q.~Huang, X.~Cao, and C.~Hou, ``Co-saliency detection
  for rgbd images based on multi-constraint feature matching and cross label
  propagation,'' \emph{IEEE Trans. on Image Process. (TIP)}, vol.~27, no.~2,
  pp. 568--579, 2017.

\bibitem{huo2018semisupervised}
S.~Huo, Y.~Zhou, W.~Xiang, and S.-Y. Kung, ``Semisupervised learning based on a
  novel iterative optimization model for saliency detection,'' \emph{IEEE
  Trans. on Neural Netw. Learn. Syst. (TNNLS)}, vol.~30, no.~1, pp. 225--241,
  2018.

\bibitem{wei2012geodesic}
Y.~Wei, F.~Wen, W.~Zhu, and J.~Sun, ``Geodesic saliency using background
  priors,'' in \emph{Proc. IEEE Eur. Conf. Comput. Vis. (ECCV)}, 2012, pp.
  29--42.

\bibitem{cheng2014global}
M.~Cheng, N.~J. Mitra, X.~Huang, P.~H. Torr, and S.~Hu, ``Global contrast based
  salient region detection,'' \emph{IEEE Trans. on Pattern Anal. Mach. Intell.
  (TPAMI)}, vol.~37, no.~3, pp. 569--582, 2014.

\bibitem{wang2015deep}
L.~Wang, H.~Lu, X.~Ruan, and M.-H. Yang, ``Deep networks for saliency detection
  via local estimation and global search,'' in \emph{Proc. IEEE Conf. Comput.
  Vis. Pattern Recognit. (CVPR)}, 2015, pp. 3183--3192.

\bibitem{lee2016deep}
G.~Lee, Y.~Tai, and J.~Kim, ``Deep saliency with encoded low level distance map
  and high level features,'' in \emph{Proc. IEEE Conf. Comput. Vis. Pattern
  Recognit. (CVPR)}, 2016, pp. 660--668.

\bibitem{li2018contrast}
G.~Li and Y.~Yu, ``Contrast-oriented deep neural networks for salient object
  detection,'' \emph{IEEE Trans. on Neural Netw. Learn. Syst. (TNNLS)},
  vol.~29, no.~12, pp. 6038--6051, 2018.

\bibitem{DSS}
Q.~Hou, M.~Cheng, X.~Hu, A.~Borji, Z.~Tu, and P.~Torr, ``Deeply supervised
  salient object detection with short connections,'' \emph{IEEE Trans. on
  Pattern Anal. Mach. Intell. (TPAMI)}, vol.~41, no.~4, pp. 815--828, 2019.

\bibitem{zhang2017amulet}
P.~Zhang, D.~Wang, H.~Lu, H.~Wang, and X.~Ruan, ``Amulet: Aggregating
  multi-level convolutional features for salient object detection,'' in
  \emph{Proc. IEEE Int. Conf. Comput. Vis. (ICCV)}, 2017, pp. 202--211.

\bibitem{li2019video}
H.~Li, G.~Chen, G.~Li, and Y.~Yu, ``Motion guided attention for video salient
  object detection,'' in \emph{Proc. IEEE Int. Conf. Comput. Vis. (ICCV)},
  2019, pp. 7274--7283.

\bibitem{EGNet}
J.~Zhao, J.~Liu, D.~Fan, Y.~Cao, J.~Yang, and M.~Cheng., ``Motion guided
  attention for video salient object detection,'' in \emph{Proc. IEEE Int.
  Conf. Comput. Vis. (ICCV)}, 2019, pp. 8779--8788.

\bibitem{CVPR_F2016}
D.~Feng, N.~Barnes, S.~You, and C.~McCarthy, ``Local background enclosure for
  rgb-d salient object detection,'' in \emph{Proc. IEEE Conf. Comput. Vis.
  Pattern Recognit. (CVPR)}, 2016, pp. 2343--2350.

\bibitem{TIP_Q2017}
L.~Qu, S.~He, J.~Zhang, J.~Tian, Y.~Tang, and Q.~Yang, ``Rgbd salient object
  detection via deep fusion,'' \emph{IEEE Trans. on Image Process. (TIP)},
  vol.~26, no.~5, pp. 2274--2285, 2017.

\bibitem{Zhu2018PDNet}
C.~Zhu, X.~Cai, K.~Huang, T.~H. Li, and G.~Li, ``Pdnet: Prior-model guided
  depth-enhanced network for salient object detection,'' in \emph{Proc. IEEE
  Int. Conf. on Multimedia and Expo (ICME)}, 2019, pp. 199--204.

\bibitem{chen2019three}
H.~Chen and Y.~Li, ``Three-stream attention-aware network for rgb-d salient
  object detection,'' \emph{IEEE Trans. on Image Process. (TIP)}, vol.~28,
  no.~6, pp. 2825--2835, 2019.

\bibitem{zhao2019contrast}
J.~Zhao, Y.~Cao, D.~Fan, M.~Cheng, X.~Li, and L.~Zhang, ``Contrast prior and
  fluid pyramid integration for rgbd salient object detection,'' in \emph{Proc.
  IEEE Conf. Comput. Vis. Pattern Recognit. (CVPR)}, 2019.

\bibitem{ronneberger2015u}
O.~Ronneberger, P.~Fischer, and T.~Brox, ``U-net: Convolutional networks for
  biomedical image segmentation,'' in \emph{International Conference on Medical
  Image Computing and Computer Assisted Intervention (MICCAI)}, 2015, pp.
  234--241.

\bibitem{simonyan2014very}
K.~Simonyan and A.~Zisserman, ``Very deep convolutional networks for
  large-scale image recognition,'' in \emph{International Conference on
  Learning Representations}, 2015.

\bibitem{he2016deep}
K.~He, X.~Zhang, S.~Ren, and J.~Sun, ``Deep residual learning for image
  recognition,'' in \emph{Proc. IEEE Conf. Comput. Vis. Pattern Recognit.
  (CVPR)}, 2016, pp. 770--778.

\bibitem{gao2019res2net}
S.~Gao, M.~Cheng, K.~Zhao, X.~Zhang, M.~Yang, and P.~H.~S. Torr, ``Res2net: A
  new multi-scale backbone architecture,'' \emph{IEEE Trans. on Pattern Anal.
  Mach. Intell. (TPAMI)}, pp. 1--1, 2020.

\bibitem{ICIP_J2014}
R.~Ju, Y.~Liu, T.~Ren, L.~Ge, and G.~Wu, ``Depth-aware salient object detection
  using anisotropic center-surround difference,'' \emph{Signal Processing:
  Image Communication}, vol.~38, pp. 115--126, 2015.

\bibitem{ECCV_P2014}
H.~Peng, B.~Li, W.~Xiong, W.~Hu, and R.~Ji, ``Rgbd salient object detection: a
  benchmark and algorithms,'' in \emph{Proc. IEEE Eur. Conf. Comput. Vis.
  (ECCV)}, 2014, pp. 92--109.

\bibitem{SSB}
Y.~Niu, Y.~Geng, X.~Li, and F.~Liu, ``Leveraging stereopsis for saliency
  analysis,'' in \emph{Proc. IEEE Conf. Comput. Vis. Pattern Recognit. (CVPR)},
  2012, pp. 454--461.

\bibitem{DES}
Y.~Cheng, H.~Fu, X.~Wei, J.~Xiao, and X.~Cao, ``Depth enhanced saliency
  detection method,'' in \emph{Proceedings of International Conference on
  Internet Multimedia Computing and Service}, 2014, p.~23.

\bibitem{LFSD}
N.~Li, J.~Ye, Y.~Ji, H.~Ling, and J.~Yu, ``Saliency detection on light field,''
  in \emph{Proc. IEEE Conf. Comput. Vis. Pattern Recognit. (CVPR)}, 2014, pp.
  2806--2813.

\bibitem{fan2017structure}
D.~Fan, M.~Cheng, Y.~Liu, T.~Li, and A.~Borji, ``Structure-measure: A new way
  to evaluate foreground maps,'' in \emph{Proc. IEEE Int. Conf. Comput. Vis.
  (ICCV)}, 2017, pp. 4548--4557.

\bibitem{Fan2018Enhanced}
D.~Fan, C.~Gong, Y.~Cao, B.~Ren, M.~Cheng, and A.~Borji, ``Enhanced-alignment
  measure for binary foreground map evaluation,'' 2018.

\bibitem{achanta2009frequency}
R.~Achanta, S.~Hemami, F.~Estrada, and S.~S{\"u}sstrunk, ``Frequency-tuned
  salient region detection,'' in \emph{Proc. IEEE Conf. Comput. Vis. Pattern
  Recognit. (CVPR)}, 2009, pp. 1597--1604.

\bibitem{liu2010learning}
T.~Liu, Z.~Yuan, J.~Sun, J.~Wang, N.~Zheng, X.~Tang, and H.-Y. Shum, ``Learning
  to detect a salient object,'' \emph{IEEE Trans. on Pattern Anal. Mach.
  Intell. (TPAMI)}, vol.~33, no.~2, pp. 353--367, 2010.

\bibitem{wang2017learning}
L.~Wang, H.~Lu, Y.~Wang, M.~Feng, D.~Wang, B.~Yin, and X.~Ruan, ``Learning to
  detect salient objects with image-level supervision,'' in \emph{Proc. IEEE
  Conf. Comput. Vis. Pattern Recognit. (CVPR)}, 2017, pp. 136--145.

\bibitem{CPD}
Z.~Wu, L.~Su, and Q.~Huang, ``Cascaded partial decoder for fast and accurate
  salient object detection,'' in \emph{Proc. IEEE Conf. Comput. Vis. Pattern
  Recognit. (CVPR)}, 2019, pp. 3907--3916.

\bibitem{BASNet}
X.~Qin, Z.~Zhang, C.~Huang, C.~Gao, M.~Dehghan, and M.~Jagersand, ``Basnet:
  Boundary-aware salient object detection,'' in \emph{Proc. IEEE Conf. Comput.
  Vis. Pattern Recognit. (CVPR)}, 2019.

\bibitem{song2017depth}
H.~Song, Z.~Liu, H.~Du, G.~Sun, O.~Le~Meur, and T.~Ren, ``Depth-aware salient
  object detection and segmentation via multiscale discriminative saliency
  fusion and bootstrap learning,'' \emph{IEEE Trans. on Image Process. (TIP)},
  vol.~26, no.~9, pp. 4204--4216, 2017.

\bibitem{chen2019multi}
H.~Chen, Y.~Li, and D.~Su, ``Multi-modal fusion network with multi-scale
  multi-path and cross-modal interactions for rgb-d salient object detection,''
  \emph{Proc. IEEE Conf. Comput. Vis. Pattern Recognit. (CVPR)}, vol.~86, pp.
  376--385, 2019.

\end{thebibliography}

\end{document}